\documentclass[sigconf]{acmart}
\makeatletter
\def\@ACM@checkaffil{
    \if@ACM@instpresent\else
    \ClassWarningNoLine{\@classname}{No institution present for an affiliation}%
    \fi
    \if@ACM@citypresent\else
    \ClassWarningNoLine{\@classname}{No city present for an affiliation}%
    \fi
    \if@ACM@countrypresent\else
        \ClassWarningNoLine{\@classname}{No country present for an affiliation}%
    \fi
}
\makeatother

\usepackage{xcolor}
\usepackage{booktabs}
\usepackage[inline]{enumitem}
\usepackage{comment}
\usepackage{ifthen}
\usepackage{algorithm}
\usepackage{algpseudocode}
\usepackage{graphicx}
\usepackage{multirow}
\usepackage{subcaption}
\usepackage{indentfirst}
\usepackage{tikz}
\usepackage{amsmath}
\usepackage{bbding}
\usepackage{pifont}
\usepackage{wasysym}
\usepackage{xltabular}
\usepackage{listings}
\usepackage{xcolor}
\usepackage{paralist}
\usepackage{wrapfig}
\usepackage{setspace}
\usepackage{amsmath}
\usepackage{blindtext}

\definecolor{codegreen}{rgb}{0,0.6,0}
\definecolor{codegray}{rgb}{0.5,0.5,0.5}
\definecolor{codepurple}{rgb}{0.58,0,0.82}
\definecolor{backcolour}{rgb}{0.95,0.95,0.92}

\lstdefinestyle{mystyle}{ 
    commentstyle=\color{codegreen},
    keywordstyle=\color{magenta},
    numberstyle=\tiny\color{codegray},
    stringstyle=\color{codepurple},
    basicstyle=\ttfamily\footnotesize,
    breakatwhitespace=false,         
    breaklines=true,                 
    captionpos=b,                    
    keepspaces=true,                 
    numbers=left,                    
    numbersep=5pt,                  
    showspaces=false,                
    showstringspaces=false,
    showtabs=false,                  
    tabsize=2
}

\AtBeginDocument{%
  \providecommand\BibTeX{{%
    Bib\TeX}}}

\setcopyright{acmcopyright}
\copyrightyear{2023}
\acmYear{2023}
\acmDOI{XXXXXXX.XXXXXXX}

\acmConference[]{}

\acmBooktitle{} 
\acmPrice{}
\acmISBN{}

\usepackage{verbatimbox}
\usepackage{fancyvrb}

\usepackage{booktabs} 
\usepackage{array}
\usepackage{verbatim} 
\usepackage{enumitem}

\usepackage{bbm}
\usepackage{caption}
\usepackage{subcaption}

\DeclareMathOperator*{\minimize}{minimize}

\usepackage{varioref}
\usepackage{cleveref}
\usepackage{xltabular}
\usepackage{xcolor}
\usepackage{paralist}

\usepackage{mathrsfs}
\usepackage{bbm}

\usepackage{bm}
\usepackage{mdframed}
\usepackage{algorithm}
\usepackage{algpseudocode}
\usepackage{algorithmicx}
\usepackage{algpseudocode}

\def\BibTeX{{\rm B\kern-.05em{\sc i\kern-.025em b}\kern-.08em
    T\kern-.1667em\lower.7ex\hbox{E}\kern-.125emX}}

\usepackage{etoolbox}
\setcopyright{none}
\renewcommand\footnotetextcopyrightpermission[1]{} 

\begin{document}

\title{MP-SL: Multihop Parallel Split Learning}

\author{Joana Tirana}
\email{joana.tirana@ucdconnect.ie}
\affiliation{University College Dublin, Ireland}

\author{Spyros Lalis}
\email{lalis@uth.gr}
\affiliation{University of Thessaly, Greece}

\author{Dimitris Chatzopoulos}
\email{dimitris.chatzopoulos@ucd.ie,}
\affiliation{University College Dublin, Ireland}

\begin{abstract}

Federated Learning~(FL) stands out as a widely adopted protocol facilitating the training of Machine Learning~(ML) models while maintaining decentralized data. However, challenges arise when dealing with a heterogeneous set of participating devices, causing delays in the training process, particularly among devices with limited resources. Moreover, the task of training ML models with a vast number of parameters demands computing and memory resources beyond the capabilities of small devices, such as mobile and Internet of Things~(IoT) devices. To address these issues, techniques like Parallel Split Learning~(SL) have been introduced, allowing multiple resource-constrained devices to actively participate in collaborative training processes with assistance from resourceful \textit{compute nodes}. Nonetheless, a drawback of Parallel SL is the substantial memory allocation required at the compute nodes, for instance training VGG-19 with $100$ participants needs $80$ GB. In this paper, we introduce Multihop Parallel SL~(MP-SL), a modular and extensible ML as a Service~(MLaaS) framework designed to facilitate the involvement of resource-constrained devices in collaborative and distributed ML model training. Notably, to alleviate memory demands per compute node, MP-SL supports multihop Parallel SL-based training. This involves splitting the model into multiple parts and utilizing multiple compute nodes in a pipelined manner. Extensive experimentation validates MP-SL's capability to handle system heterogeneity, demonstrating that the multihop configuration proves more efficient than horizontally scaled one-hop Parallel SL setups, especially in scenarios involving more cost-effective compute nodes.

\end{abstract}

\maketitle
\makeatletter
\patchcmd{\maketitle}{\@copyrightspace}{}{}{}
\makeatother

\section{Introduction}
\label{sec:intro}

\begin{figure}[t]
    \centering
    \includegraphics[width=1\columnwidth]{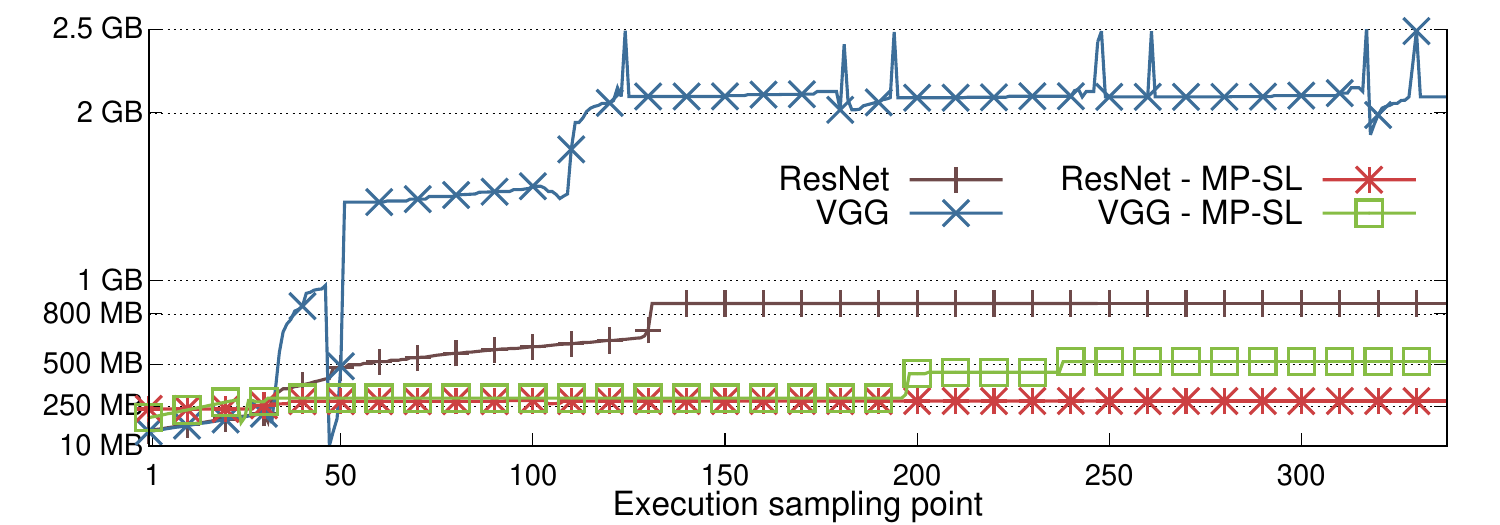}
    \caption{Memory usage 
    measured 
    on a Raspberry~Pi~4, when training ResNet-101 and VGG-19, 
    with FL and  MP-SL.}
    \label{fig:memory}
\end{figure}

Modern mobile and IoT devices feature 
different sensors that produce rich and voluminous data, which can be used to train deep Neural Network~(NN) models. Given that such devices have limited computing resources, a typical way to do this is gathering the data in one machine (\emph{server}) and training the model there. 
However, such solutions require powerful computing infrastructure on the server and are not privacy-preserving
as the users' data are exposed. Collaborative training approaches, such as Federated Learning~(FL)~\cite{mcmahan2017communication}, enable training models in a distributed manner, 
while keeping the data decentralized. 
Usually, the devices that participate in FL are referred to as \emph{data owners}. 
Data owners train the model locally (\emph{on-device training}) for a small number of rounds, and share the updated model with a server. 
The server, in turn, \emph{in an aggregation phase}, produces a global instance of the model. 
Finally, the updated global model is sent back to the data owners to start a new training epoch.

However, on-device training, even for a limited number of rounds, can be quite computationally intensive and demanding in terms of computing resources. Recently released mobile GPUs that are designed to support the training of NN models, still perform poorly~\cite{gim2022memory}. One factor is the large batch size requirement to ensure good accuracy and convergence speed~\cite{wang2022melon}. Thus, during training, large tensors containing the produced activations during forward propagation will be generated. Depending on the size of the model, this may require the device to allocate a considerable amount of memory to be able to train the model.
For example, Fig.~\ref{fig:memory} depicts the portion of memory occupied by a process that is held in the main memory during the training process in FL (data owner has the full model). 
However, in FL, this is significantly high for constrained devices to handle. For instance, there are small devices like the RPi 3, or the GPU of NVIDIA Jetson Nano, that cannot support such type of training. Nevertheless, even if a device can perform on-device training, its limited computing resources will cause the \emph{stragglers effect} whereby slower devices can lead
to unacceptably large training delays.

Typically, most FL applications consider models with fewer parameters, like the shallower 
versions of ResNet and VGG~(e.g., ResNet18, VGG11,~etc.).
But, in Split Learning protocols (SL)~\cite{vepakomma2018split,10.1145/3446382.3448362} 
the largest part of the model is assigned to a \textit{compute node} that performs the respective training process, whereas data owners only keep a small part of the model. This makes it possible for resource-constrained data owners to train deeper models. Moreover, SL is beneficial in terms of
communication load. Namely, data owners need to exchange only the intermediate activations and gradients with the compute node, 
whereas in FL they have to exchange the parameters of the entire model with the aggregator -- this can be hundreds of MB.

\begin{figure*}[h]
\centering
\begin{subfigure}[b]{0.5\columnwidth}
    \centering
    \includegraphics[clip,width=\columnwidth]{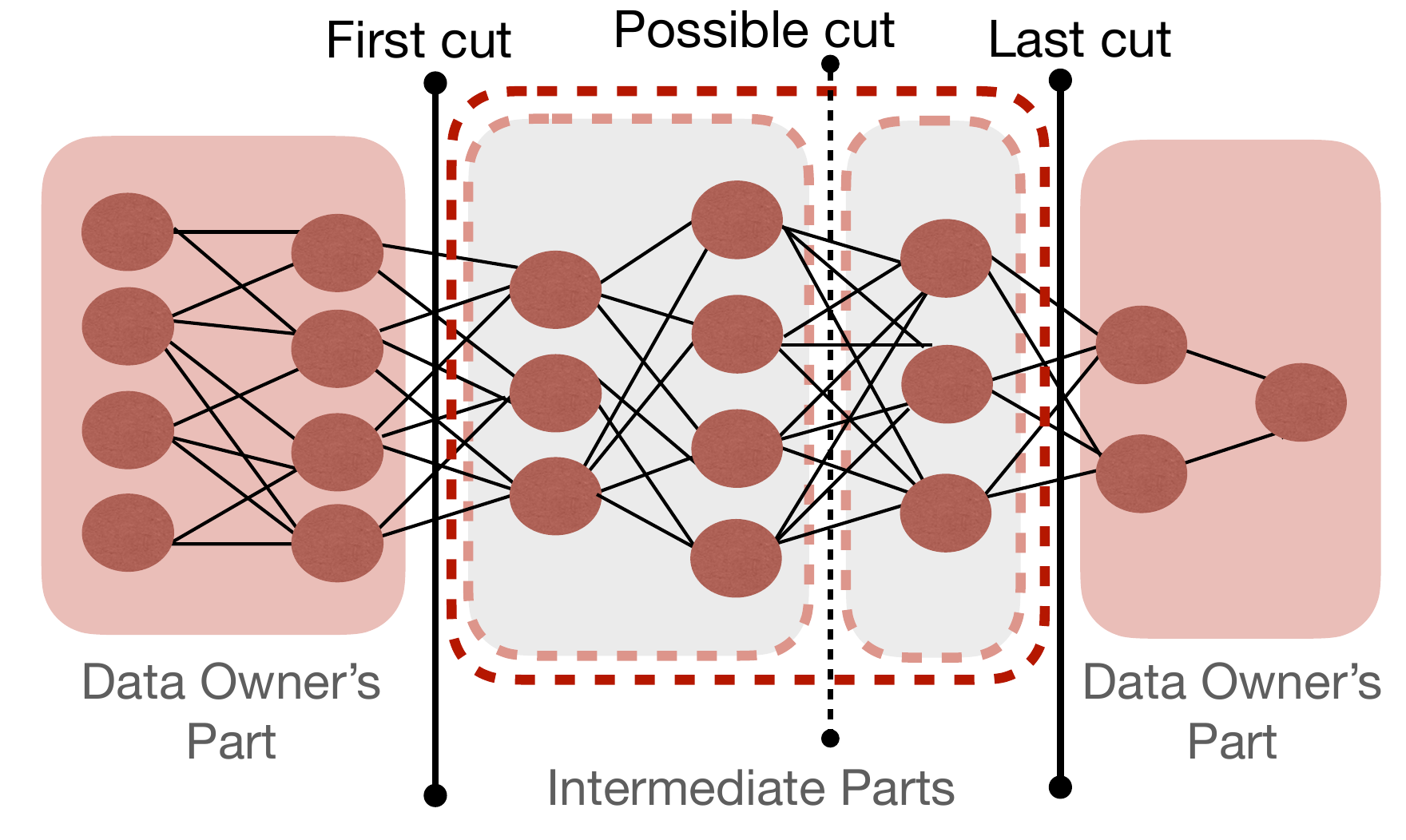}
    \caption{Model partitioning.}
    \label{fig:model_seg}
\end{subfigure}
\begin{subfigure}[b]{0.8\columnwidth}
    \centering
    \includegraphics[clip,width=\columnwidth]{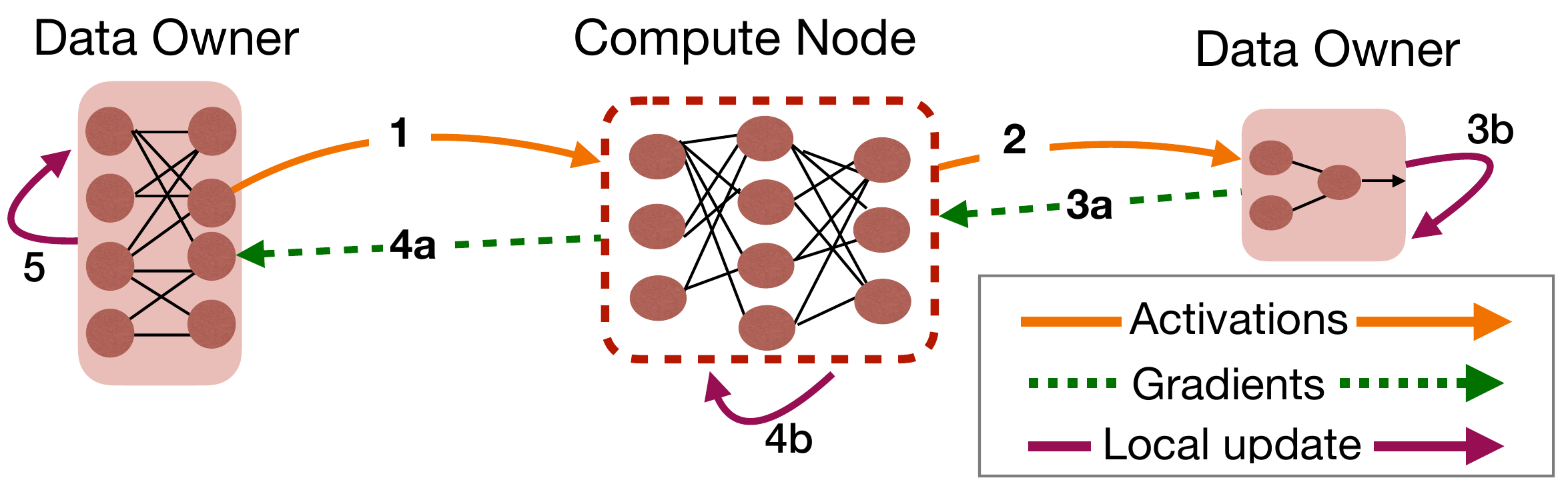}
    \caption{One batch update in SL with one compute node.}
    \label{fig:sl_workflow}
\end{subfigure}
\begin{subfigure}[b]{0.7\columnwidth}
    \centering
    \includegraphics[clip,width=\columnwidth]{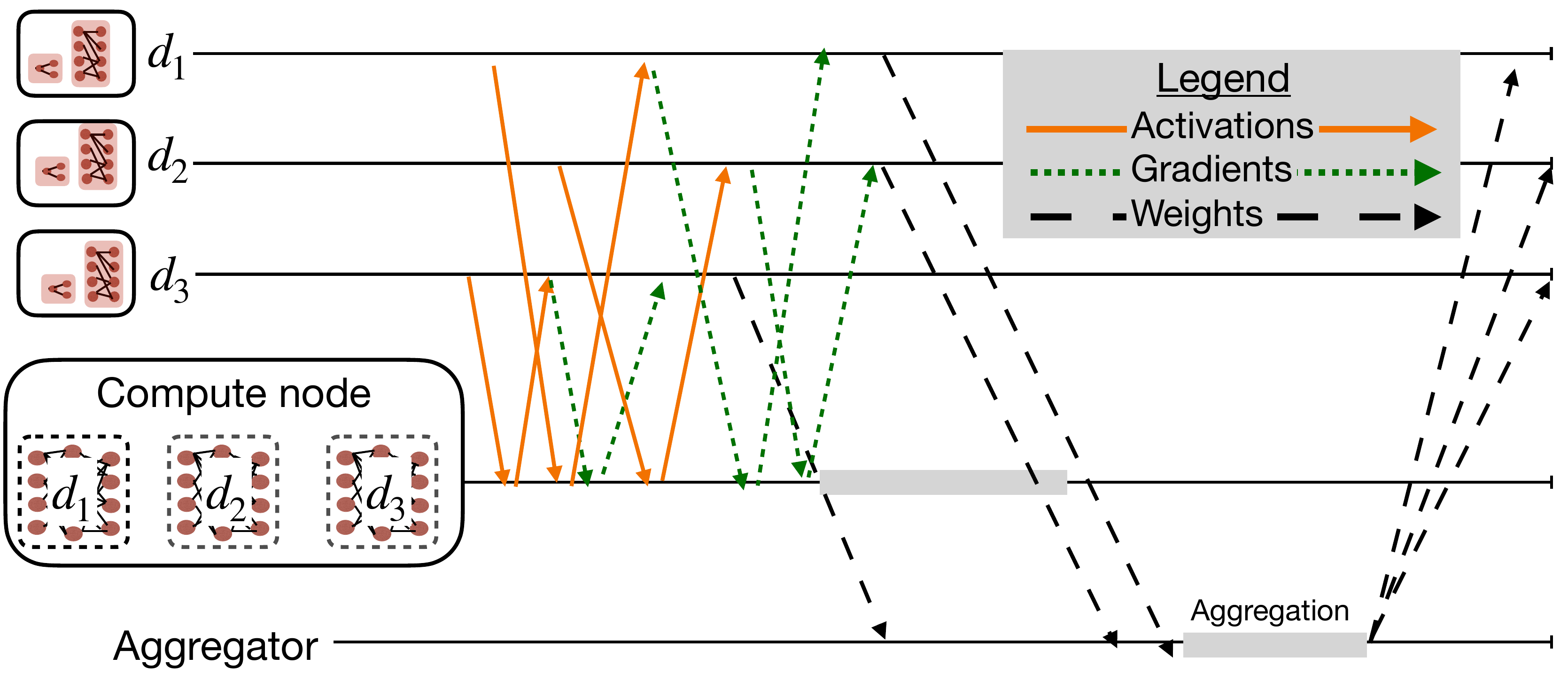}
    \caption{Parallel SL protocol with three data owners.}
    \label{fig:parallel_split}
\end{subfigure}
\caption{Applying SL requires (a)~model split and (b)~communication with at least one compute node. When more than one data owners participate (c)~Parallel SL can ensure scalability by allowing data owners to make model updates independently.\vspace{-0.1cm}}
\label{fig:sl_}
\end{figure*}

In this work, we present Multihop Parallel SL~(MP-SL), a distributed learning framework that combines SL with FL in a way that achieves better scalability. Note that MP-SL is an orthogonal approach to FL since it enables resource-constraints devices to participate in the training of large deep-learning models and restrains the stragglers effect. This is presented in Fig.~\ref{fig:memory}, where the on-device memory demands can be dropped 
 up to $76\%$ with MP-SL. Furthermore, MP-SL supports model splitting into multiple parts that are assigned in different compute nodes~\cite{tirana2022role} 
to \textit{(i)}~relax the memory requirements of each compute node, \textit{(ii)}~reduce the cost of compute nodes,
and \textit{(iii)}~restrict the model's exposure. 
Data owners can choose their desired 
multihop level, which internally translates to a suitable partitioning of the model, with each part being assigned to a different compute node allowing pipelined parallelism. In fact, given the desired multihop level, MP-SL will optimize the selection of the intermediate split points~(i.e., the model parts assigned to the compute nodes) by minimizing the training latency. 

\noindent In summary, our contributions are: 

\smallskip  \noindent \textbf{1.} MP-SL is the first Parallel SL-based framework with multihop support. It is modular, easily extensible to support any model type,
and is publicly available.~\footnote{\url{https://github.com/jtirana98/MultiHop-Federeated-Split-Learning}}

\smallskip  \noindent \textbf{2.} We provide and validate an analytical model for estimating the expected performance of MP-SL. We show that the analytical model provides estimates of the measured system performance, 
with an error less than $3.86\%$. 

\smallskip  \noindent \textbf{3.} To the best of our knowledge this is the first work that models and optimizes the splitting selection for multihop SL.

\smallskip  \noindent \textbf{4.} We evaluate MP-SL for a wide range of scenarios using a realistic testbed. We show that the proposed protocol is robust to the stragglers effect and can significantly reduce the cost of the compute nodes with a slight increase in the training time.

\section{Background and Related Work}
\label{sec:background}

\subsection{Parallel Split Learning} Typically in SL protocols, the model  
is vertically split into multiple parts, with a subset of them being offloaded to more powerful compute nodes. 
Fig.~\ref{fig:model_seg} depicts a model that is split into different parts through \textit{cut layers}. 
The first and the last part of the model~(before the first cut and after the last cut, respectively) are kept locally at the data owner, while the intermediate part is assigned to a compute node or further divided~(via the possible cut) and assigned to two compute~nodes. 

Moreover, Fig.~\ref{fig:sl_workflow} 
shows one training iteration between one data owner and one compute node~(single hop). The data owner initiates each iteration. Firstly, during the forward-propagation, the nodes send to each other~(starting from the first model part) \verb|forward()| requests containing the activations produced at the corresponding cut layers~(steps 1-2). Then, during the back-propagation, following the opposite direction, and starting from the last model part the nodes compute the gradients and encapsulate the ones of the cut layers into \verb|backward()| requests~(steps 3a, 4a). When a node has computed the gradients, it can concurrently update the weights of the model it is in charge of~(steps 3b, 4b).

For multiple data owners in the conventional SL approach~(i.e, SplitNN~\cite{vepakomma2018split}), they share a common instance of the intermediate model part and are serviced by the compute node in a round-robin fashion. However, the sequential serving of the data owners increases the training delay. 
\textit{Parallel SL}~\cite{thapa2022splitfed,jeon2020privacy,wu2023split}, speeds up training and enhances scalability. As is shown in Fig.~\ref{fig:parallel_split}, the compute nodes keep a different version of the model parts for each data owner. This allows each data owner to apply SL independently from  
other data owners. At the end of an epoch,  
the model parts from all the data owners 
are \textit{aggregated} using techniques such as FedAvg~\cite{mcmahan2017communication}, in which the data owners employ an aggregator~\cite{thapa2022splitfed}. 
Notably, the intermediate model parts can be aggregated locally at the compute nodes, without any communication.

Parallel SL performance is constrained by compute nodes' capacity, especially for numerous data owners. A relatively straightforward way to scale for a large number of data owners is  
to apply \textit{horizontal scaling}, involving the addition of more compute nodes that 
can serve different sets of data owners. However, this introduces synchronization challenges for completing epochs, as compute nodes must aggregate intermediate model parts. Also,  
in Parallel SL, each compute node has to hold the entire intermediate part of the model for each data owner they are in charge of. 
This becomes problematic for models with extensive parameters, demanding substantial memory, and often requiring powerful (and costly) compute nodes.

\begin{figure}[t]
    \centering
    \includegraphics[width=1\columnwidth]{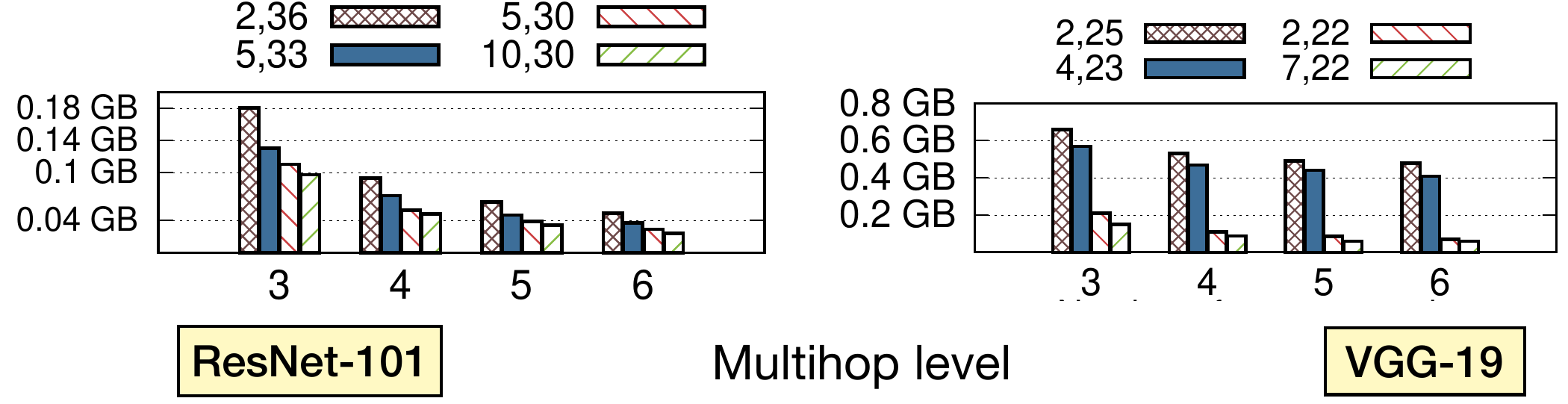}
    \caption{Memory demand for the compute node with the largest (memory-wise) model part for different multihop levels. The smallest multihop level is $3$~(i.e., one compute node), and the largest is $6$~(i.e., four compute nodes). Also, for each model, we select different user-defined first and last cut layers. Note that VGG19 has $25$ indivisible layers while ResNet101 has~$37$. \vspace{-0.9cm}}
    \label{fig:mem_reduction}
\end{figure}

\subsection{Multihop Parallel SL} Going a step further, the intermediate model part, can be split into smaller parts, which can be assigned to separate compute nodes. Specifically, by splitting further the 
model parts, we observe the following advantages: 

\smallskip \noindent \textit{\underline{(1) Resource relaxation:}}~Memory and processing requirements for compute nodes are relaxed, enabling the use of less powerful and more affordable compute nodes. 
Fig.~\ref{fig:mem_reduction} shows the reduction of the memory demands on the compute nodes while the multihop level increases. In Fig.~\ref{fig:mem_reduction}, the first/last cut layers are user-defined. While the intermediate model parts are calculated using MP-SL which, finds the split policy that minimizes the training delay~(see Sec.~\ref{sec:split}). Also, according to AWS pricing, the expense of renting Virtual Machines (VMs) can notably decrease through the reduction of memory capacity. For example, the instance type t2.large~(8 GB memory, 2 vCPUs) costs 0.0928 USD/h, while t2.medium~(2 GB, 2 vCPUs) costs 0.0464 USD/h. Hence, we can rent a VM with the same computing capacity but at almost half the price. 

\smallskip \noindent \textit{\underline{(2) Knowledge conceal:}}~As the multihop level increases, each compute node is in charge of smaller model parts~(i.e, consisting of a smaller number of layers) and hence has fewer knowledge about the ML model, which is an essential aspect in SL. 
The research area which focuses on privacy concerns of SL, mostly involves semi-honest attacks with a single split~\cite{li2021label, liu2023distance}. In contrast, in our case we build upon the no-label sharing configuration~(i.e, at least two splits). 
Also, the attacks  
depend on the received activations/gradients, which 
can be protected with defense techniques~\cite{vepakomma2020nopeek}. Only the compute nodes of PCAT~\cite{gao2023pcat} exploit the model part they are in charge of. But, even though PCAT outperforms other novel attacks, it remains sensitive to the number of offloaded layers, and hence one can challenge PCAT by increasing the multihop level.

\smallskip \noindent \textit{\underline{(3) Pipeline Parallelism~(PP):}} Splitting the model into multiple parts, enables
PP~\cite{narayanan2019pipedream}, which is the combination of Data Parallelism~(DP)~\cite{cui2016geeps} and  Model Parallelism~(MP)~\cite{huang2019gpipe}. The benefits of PP are (i)~accelerated training and (ii)~support for larger models.
However, it is mainly used in centralized learning for a single source of data~(i.e., one data owner). 
Even though DP and MP have been adopted by decentralized learning with SplitFed~\cite{thapa2022splitfed} and other variants of Parallel SL~\cite{wu2023split}, the concept of PP has not been widely explored in such a configuration. However, in this case, the implementation of PP is even more challenging as (i)~the compute nodes store multiple instances of the model~(one for each data owner), (ii)~the first/last model parts are handled by resource-constrained devices, and (iii)~the compute nodes do not have access to the data. Therefore, existing PP techniques~\cite{jia2019beyond,zheng2022alpa} are not applicable.

Nevertheless the benefits of multihop SL, there are not many examples. For instance, in CHEESE~\cite{cheng2023cheese} the 
nodes form clusters to help each other. In each cluster, the model is split into a number of parts equal to the number of nodes inside it.
The communication between the nodes relies on a device-to-device system. However, it is known that such systems are not fully implemented~\cite{asadi2014survey}, yet. Also, each cluster helps only one data owner; hence, it does not consider PP for additional acceleration. 
Similarly, FedSL~\cite{abedi2020fedsl}, supports multihop, but
exclusively focuses on implementing recurrent NNs.

\subsection{Cut layer selection} In SL, one of the most crucial decisions is identifying the cut layers since they determine the model parts, consequently affecting computing and communication delays per node. In existing research work, the most commonplace considered system is the one with multiple data owners and one compute node~(i.e., one-hop)~\cite{wu2023split,kim2023bargaining,samikwa2022ares}. Typically, in these works the approach is to build a mathematical model of the system and then optimize its parameters~(e.g., energy consumption, delay, etc.).
Alternatively, \cite{wu2022fedadapt} uses Reinforcement Learning~(RL) to find the best split. But, as the system scales more RL agents need to be used. Also one should consider the overhead of training the RL agent. Only CoopFl~\cite{wang2023coopfl} and \cite{tirana2024workflow} consider the case of multiple compute nodes but only in the horizontally scaled Parallel SL context~(not in the multihop configuration).

\subsection{Machine Learning as a Service~(MLaaS)} MP-SL framework provides the user an MLaaS functionality, that implements a collaborating learning protocol with offloading without the user's intervention.   
Many works allow this for FL protocols~\cite{ziller2021pysyft,samir2018pygrid,flower}, but little has been done in the case where offloading is essential. 
For instance, OpenMined released a blog post~\footnote{\href{https://medium.com/analytics-vidhya/split-neural-networks-on-pysyft-ed2abf6385c0}{https://medium.com/analytics-vidhya/split-neural-networks-on-pysyft-ed2abf6385c0}} that extends PySyft~\cite{ziller2021pysyft} for SL. 
However, it is an approach for one data owner with a centralized orchestration, not allowing the addition of any extra functionalities in the compute nodes. Unlike, MP-SL, in which compute nodes  
can be easily re-programmed. Another MLaaS framework is Blind Learning~\cite{gharibi2022automated} which supports SL over the internet.   
But it is a commercial product, without an open-source implementation.
\section{Design of MP-SL Framework}
\label{sec:system}

We propose MP-SL, a solution that can stand as an alternative to FL when on-device training is not fully supported by the participating devices. Therefore, we adopt the no-label sharing~\cite{vepakomma2018split} configuration of SL, which is closer to the properties of FL. Also, to allow the system to scale,  
MP-SL implements the pipelined parallel multihop SL protocol~\cite{tirana2022role}.

\begin{figure}[t]
\centering
\includegraphics[width=0.9\columnwidth]{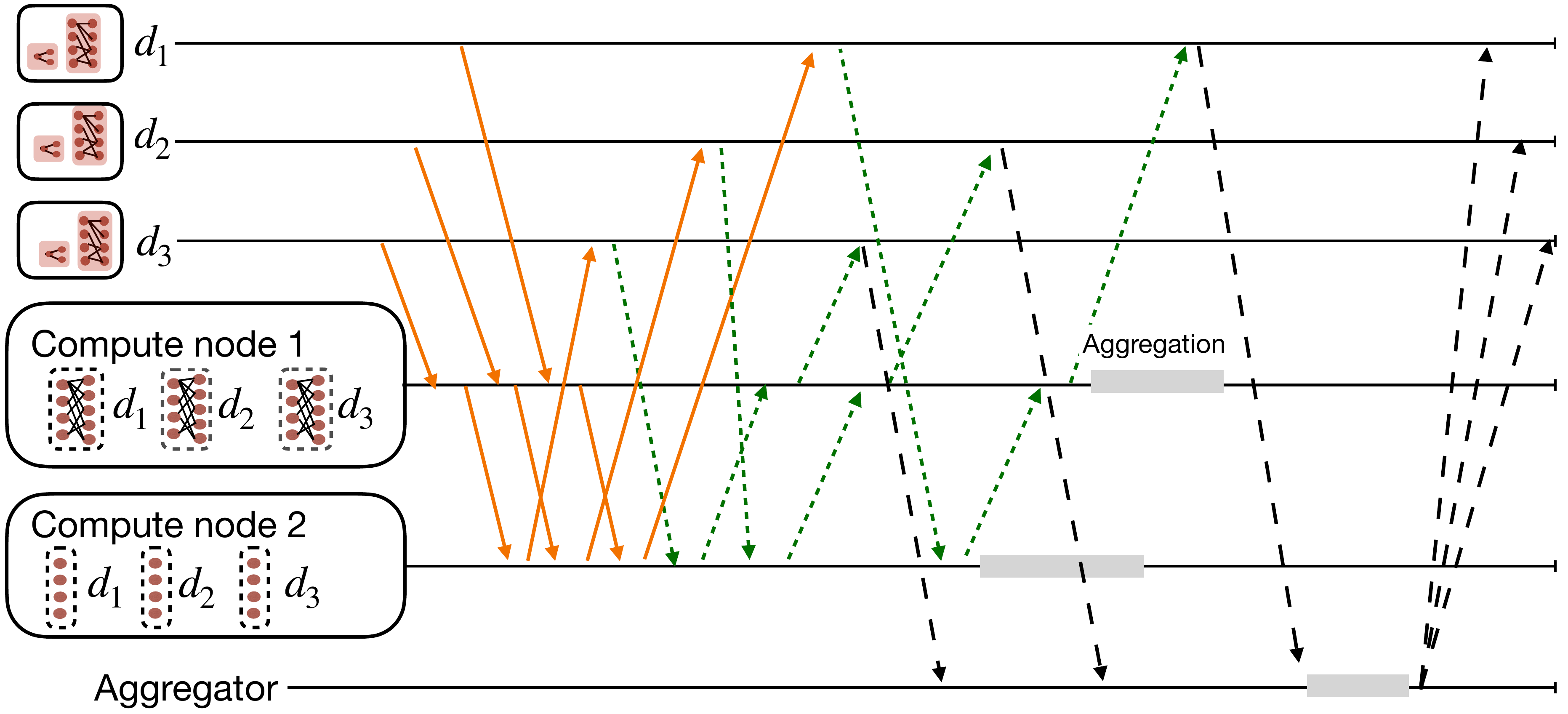}
    \caption{ MP-SL protocol with two compute nodes. 
    \vspace{-0.5cm}}
    \label{fig:MP-SL}
\end{figure}

Like in Parallel SL, the compute nodes, for each data owner, maintain a different copy of the model part they are in charge of. This will allow the data owners to apply SL asynchronously. As is shown in Fig.~\ref{fig:MP-SL} data owner~($d_1$) sends a \verb|forward()| request to compute node~1. Then, the compute node~1 will propagate the activations from the first cut layer to the model part that belongs to $d_1$ and will send a new \verb|forward()| task~(containing the activations from the successive cut layer) to compute node~2. Concurrently, data owner~($d_2$) applies SL in its model parts. 
This order follows the inference path, whereas the \verb|backward()| tasks will follow the inverse path.
Notably, this parallel execution of the \textit{sub-tasks} follows a similar execution of pipeline computing. Finally, the data owners will synchronize their states in the aggregation phase. The intermediate model parts are locally and asynchronously aggregated in each compute node, whereas the data owners send the model parts to an aggregator. Next, we discuss the design of MP-SL framework in detail. 
  
\subsection{Setting up MP-SL}

\smallskip
\noindent \textbf{Entities}. MP-SL is a distributed framework. 
The main entity is the \emph{Manager} node of the system, 
which receives a new training request and is in charge of setting up and preparing the system to execute the request. It participates in the execution as the  
aggregator for the data owner's model parts. 
Here we assume that the manager is an honest node, but the framework can be extended to use secure aggregation techniques~\cite{bonawitz2017practical}. 
One of the data owners is the \emph{init device} that will send the job request to the manager. Whereas, the manager will select a group of compute nodes to execute the training.

\smallskip
\noindent \textbf{Overview of job execution.} 
We assume that all participating nodes have already installed and compiled the source code of MP-SL, which is written in C++. The init node submits a new \emph{job} to the manager by sending a YAML file containing the description of the training task. Namely, this contains the multihop level, the first/last cut layer~(i.e, the model parts that will not be offloaded), the model's name, which corresponds to an entry inside the MP-SL's artifact and may contain some training hyper-parameters~(e.g., batch size, learning rate, etc.). 
Upon receiving a training request, the manager locates the participating data owners and compute nodes. Subsequently, it dispatches a configuration message instructing the nodes on the internal initialization of the framework's module. Then, the nodes retrieve the model-specific code from the artifact, enabling the commencement of the training phase.

\noindent \textbf{Model Submission.} The manager node holds an artifact with the models' code. 
The framework already provides the implementation of some CNN-based models, such as ResNet~\cite{he2016deep}, and VGG~\cite{simonyan2014very}. In the artifact, each model is stored in a separate directory, that contains an \emph{hpp} file of the API functions, through which one can generate the desired model parts.
The user can submit a new model entry inside the artifact, by creating a new directory with the following API calls: 

\noindent (i)~\verb|model()| describes the architecture of the model and declares the model's \emph{atomic unit-blocks}~\footnote{We use the terms atomic unit-block and layer interchangeably.} (i.e., parts of the model that cannot be split any further according to users definition). An atomic unit-block may contain only one layer or multiple sequential layers. For instance, a ResNet's \textit{resblock} is an atomic unit-block that cannot be split any further due to the complicated connections between the neurons. 

\noindent (ii)~\verb|model_part(int start, int end, ...)| function will be used at the runtime to generate a specific model part that contains the layers from the \verb|start|-point until the \verb|end|-point. It calls the \verb|model()| to learn which are the atomic unit-blocks that correspond to the requested model part. For instance, in Fig.~\ref{fig:model_seg} there are two atomic unit-blocks (grey color), hence we can define a model part starting from the \textit{first cut} until the \textit{last cut} or, in case of higher multihop level, we can split one more time using the \textit{possible cut}.

\noindent (iii)~\verb|split_rule(int multihop-level, int p)| will return the \verb|start| and \verb|end| values that the node $p$ is responsible for. 
The total number of model parts is defined from the user-defined multihop level.

\begin{figure}[t]
    \centering
    \includegraphics[width=1.\columnwidth]{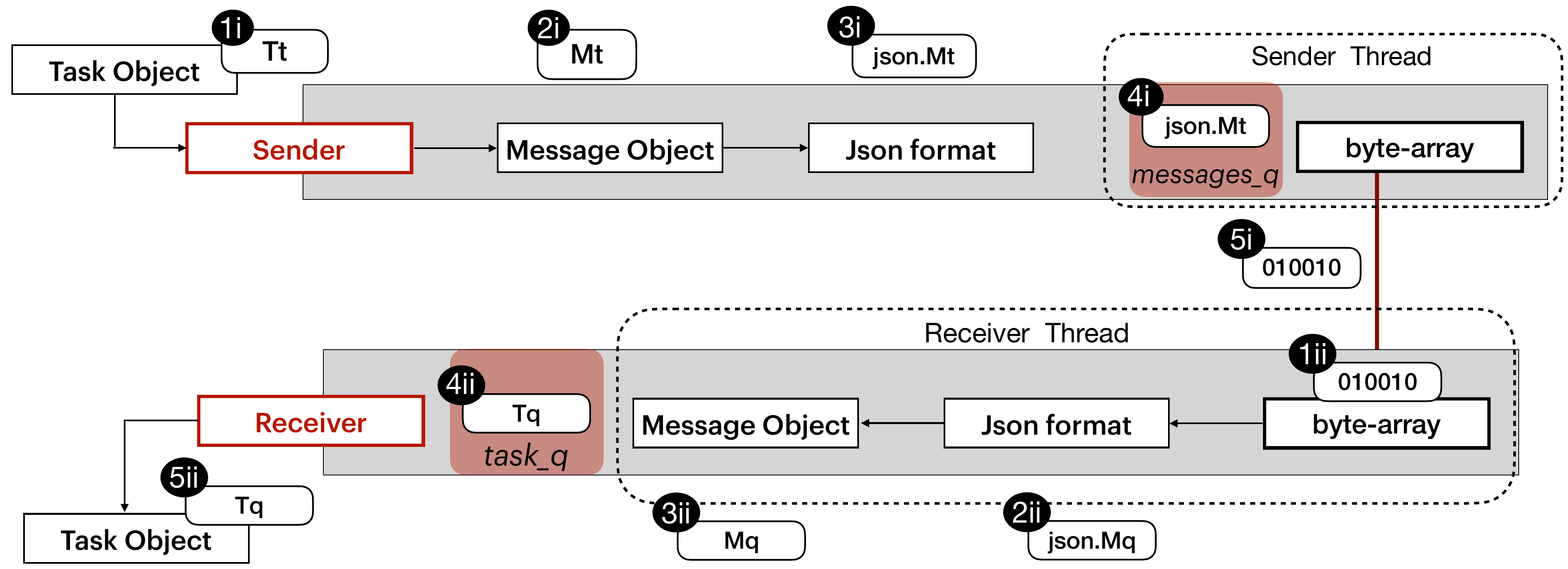}
    \caption{Task serialization and message encapsulation.
    \vspace{-0.5cm}}
    \label{fig:message}
\end{figure}

\subsection{MP-SL modules}
There are two main modules, the \emph{task delivery} module that handles the communication between the entities, and the \emph{SL engine} that implements the SL steps.

\noindent \textbf{Task Delivery.} 
It provides the following API calls:
\begingroup
\makeatletter
\@totalleftmargin=-0.4cm
   \begin{Verbatim}
    void send(task_t Tt, int node_id)
    task_t receive()
   \end{Verbatim}
\endgroup

\noindent Where \verb|send()| is a non-blocking function that receives the task description
that should be sent to \verb|node_id| (step $1i$ of Fig.~\ref{fig:message}). 
\begin{wrapfigure}[10]{l}{5cm}
\raggedleft
\vspace{-0.3cm}
    \includegraphics[width=5cm]{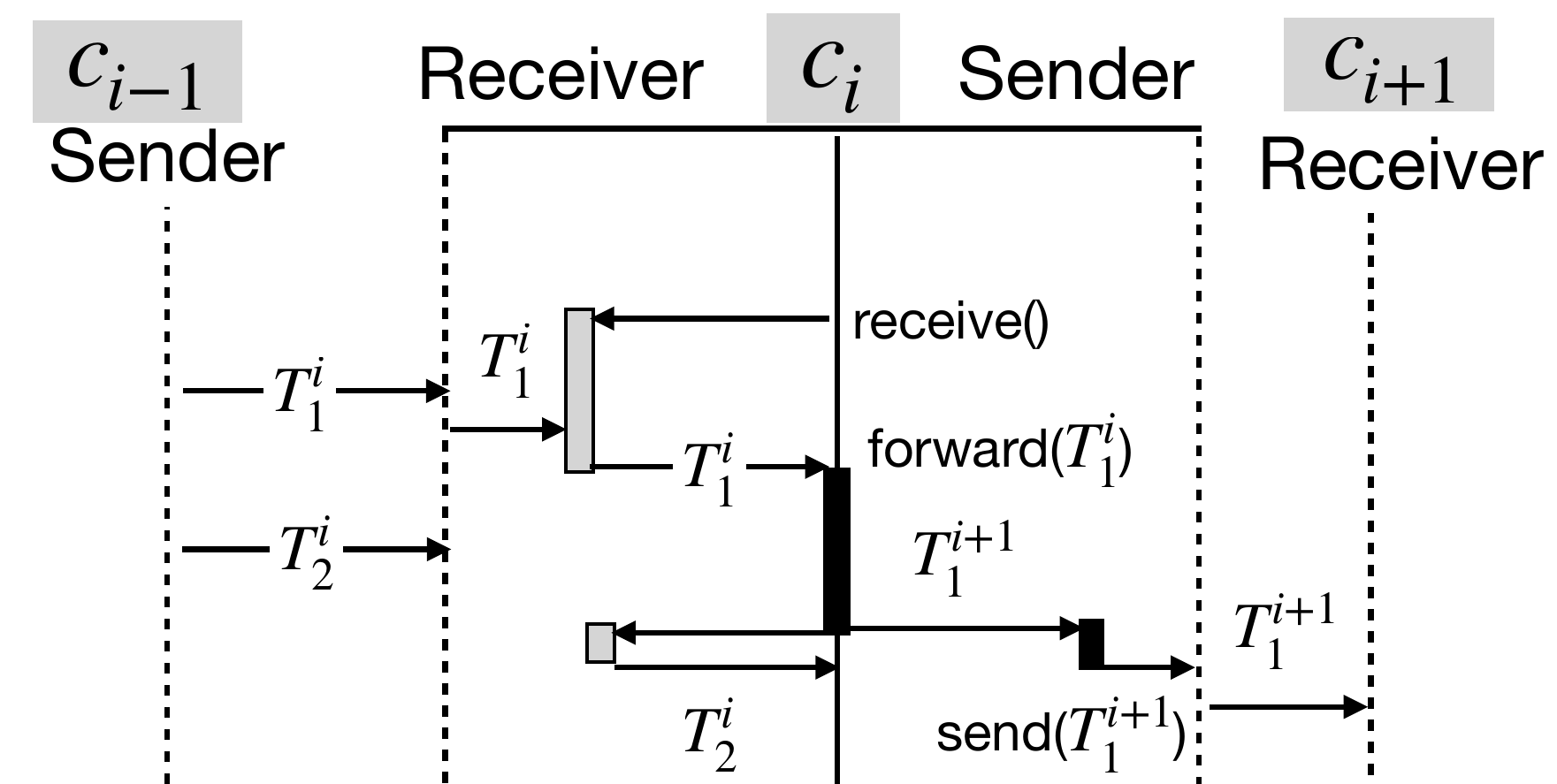}
    \caption{Function call example.
    \vspace{0.5cm}}
    \label{fig:functions}
\end{wrapfigure}
The module will generate a task-message $M_t$ from the task object by serializing it into a JSON message (steps $2i-3i$, Fig.~\ref{fig:message}). 
Then, the message will be inserted in a queue~(step $4i$). 
The \emph{sender} thread periodically checks the queue for new messages. It is responsible for encapsulating the message into a TCP packet and transmitting it to \verb|node_id|~(step $5i)$.
Respectively, when a node receives a new message-task the \emph{receiver}  thread will apply the reverse steps to decapsulate and unserialize the message into a task (steps $1ii$~to~$5ii$, Fig.~\ref{fig:message}). The received task-messages are stored in a queue inside the task-delivery module and can be retrieved through \verb|receive()|, a blocking function call that returns the first task of the queue.

Fig.~\ref{fig:functions} shows how the communication overhead is overlapped by the computing tasks. 
It contains the main and the two
background threads~(receiver and sender) for node $c_i$, the sender for $c_{i-1}$, and the receiver for $c_{i+1}$. 
The example shows instances of forward tasks, in which $c_{i-1}$ sends tasks to $c_{i}$ and then $c_{i}$ generates tasks for $c_{i+1}$. 
It illustrates how a task's computation is overlapped by another task's transmission delay. At first, $c_i$ receives $T^i_1$, then while $c_i$ is executing the forward operation (to produce the task $T_1^{i+1}$ for $c_{i+1}$) the $c_{i-1}$ is sending the successive task. So, when $c_i$ completes $T_1^i$, it will start the task $T_2^i$ immediately without any delay.

\noindent \textbf{SL Engine.} This module is in charge of executing the tasks that are generated while the training procedure takes place. 
The module first receives the training parameters (e.g., learning rate, batch size, etc.), the model name, the split points that define the model parts, and in the case of the compute node, a vector with the clients' IDs. Then, the SL engine 
generates the model parts using the \verb|model_part()|. 
For the case of the compute nodes, it creates a different copy of the model part for each client in the vector; as Parallel SL requires. The module stores the state for each model part it generates. The state contains the weights of the model part and the activations generated during the forward pass.
Also, it applies the learning tasks through the following API calls:

\begingroup
\makeatletter
\@totalleftmargin=-0.4cm
   \begin{Verbatim}
    task_t forward(task_t Tt), 
    task_t backward(task_t Tt)
   \end{Verbatim}
\endgroup

\noindent The \verb|task_T| object contains the information needed for the module to execute the task~(e.g., a tensor with activations/gradients, data owner's id, etc.). When a task execution finishes, the module generates the next task for the respective node of the inference path. 
Finally, during the aggregation phase, the module updates the model part, using the following API:

\begingroup
\makeatletter
\@totalleftmargin=-0.4cm
   \begin{Verbatim}
    void updateModel(Tensor globalModel)
   \end{Verbatim}
\endgroup

\noindent The data owners will use the \verb|globalModel| received from the aggregator. Whereas, the compute nodes can apply the aggregation locally, and thus can ignore this parameter.
\section{Training cost model}
\label{sec:model}

In this section, we capture the cost of training using the MP-SL framework via a 
model, which focuses 
on the main computing and communication dimensions and overheads of the training process. The model is 
validated in the evaluation (Section~\ref{sec:evaluation}), 
by comparing its estimates with the results obtained via real model training in a testbed. It is also used to explore what-if scenarios for configurations of larger scale. 

\smallskip \noindent \textbf{System model.}
Consider a set of data owner nodes $d_k, 1 \leq k \leq K$ who wish to collaboratively train an ML model $M$. Assume that each data owner has the same number of batches $B$.~\footnote{MP-SL supports scenarios with varying data volumes among data~owners. This assumption is made to avoid the over-complexity of the cost model.}
Let $mem_M$ represent the required memory for hosting the model on a node. Also, let $procT_M(n) = procT^{fwd}_M(n) + procT^{back}_M(n)$  
be the processing time required for the forward-and-back-propagation steps for a single batch, given the processing capacity of node $n$.

MP-SL is designed to enable data owners who may not have sufficient memory to accommodate the 
model and/or are not powerful enough to perform the respective computation in an acceptable time. To train the model,
data owners can use one or more compute nodes $c_i, 1 \leq i \leq N$, with memory $mem_{c_i}$. Data owners communicate with compute nodes over wired or wireless links while compute nodes communicate with each other typically over a fast wired network.
Let $bw_{n_i,n_j}=bw_{n_j,n_i}$ denote the bandwidth of the (symmetrical) link between nodes $n_i$ and $n_j$ (data owners or compute nodes).

\begin{figure*}[ht]
\begin{center}
\subfloat[Each task 
$\tau_x$ is split in two parts assigned to nodes $a$ and $b$.]{\includegraphics[width=0.49\textwidth]{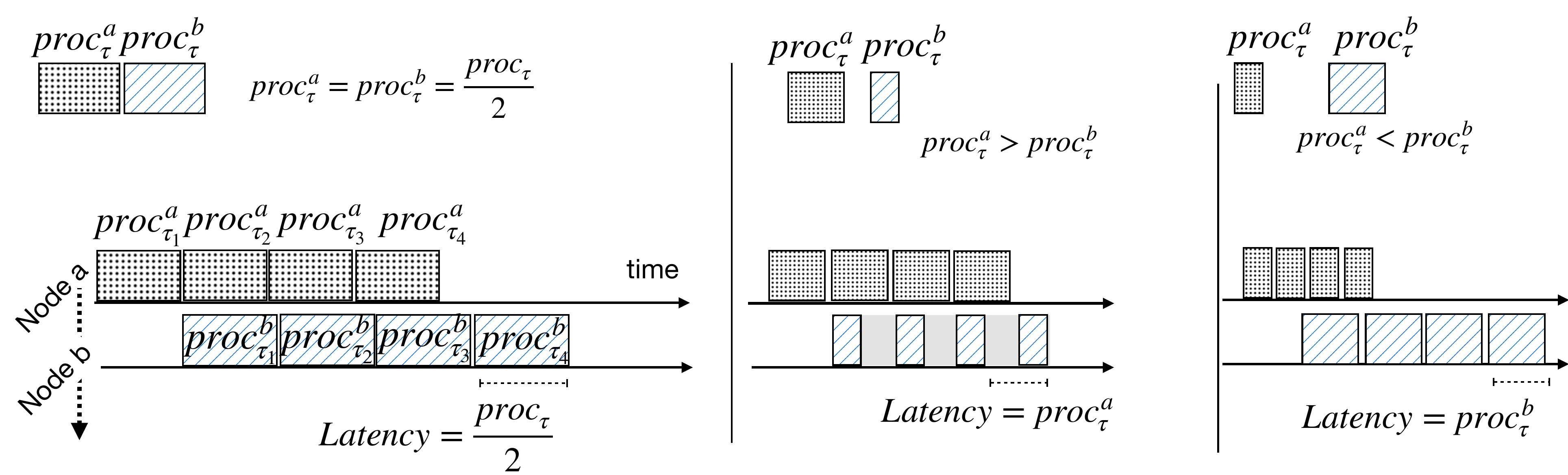}
  \label{fig:one_split}
} 
\subfloat[Each task  
$\tau_x$ is split in three parts assigned to nodes $a$, $b$ and $c$.]{\includegraphics[width=0.49\textwidth]{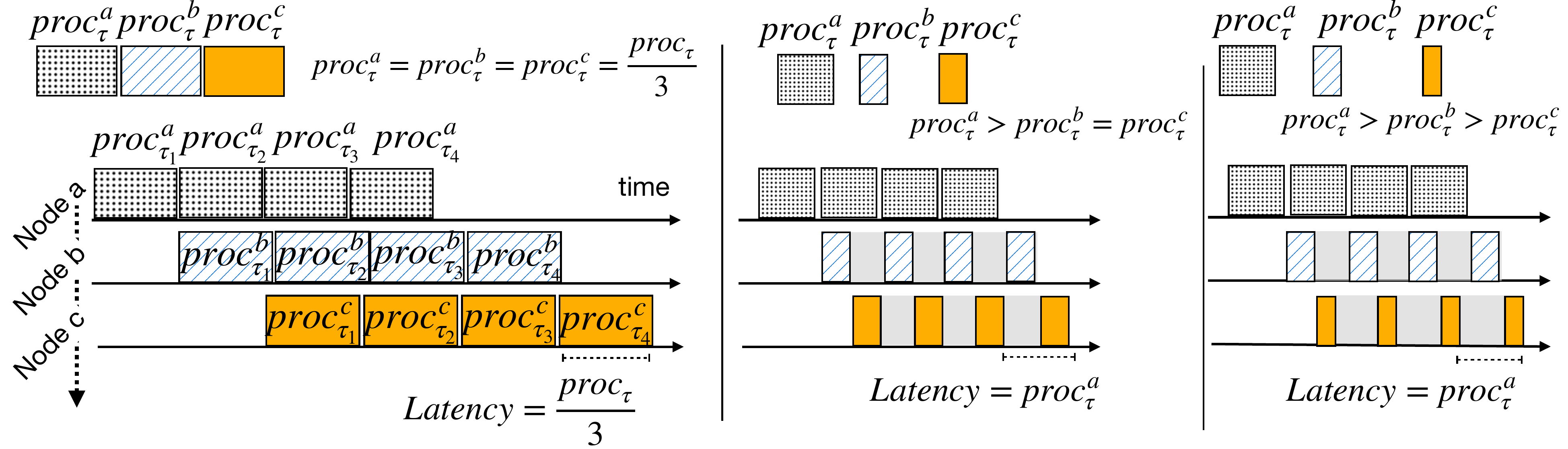}
  \label{fig:two_split}
}
\end{center}
\vspace{-0.2cm}
\caption{Pipeline latency for different task splitting scenarios. Four training tasks $\tau_1, \tau_2, \tau_3, \tau_4$ 
are split into smaller 
parts, assigned to 
(a) two or (b) three nodes in a pipeline (vertical direction). The boxes 
represent the time needed by the nodes to process their 
parts (horizontal direction). 
The grey areas denote the waiting times for the completion of the previous parts.\vspace{-0.2cm}}
\label{fig:pipeline-throughput}
\end{figure*}

\smallskip
\noindent \textbf{Model splitting.}
The model $M$, which consists of $S$ atomic unit-blocks, is split in $P$~(i.e, multihop level) parts $M_p, 1 \leq p \leq P$, with each part consisting of one or more consecutive atomic unit-blocks of the model. 
Let $mem_p < mem_M$ be the memory needed to host part $M_p$, and let $procT_p(n) < procT_M(n)$ be the processing time required to train part $M_p$ on node $n$. 
Let $n^k_p$ represent the node to which part $M_p$ is assigned for the training process of data owner $d_k$. For such an assignment to be feasible, the node must have sufficient memory to host the part, $mem_{n^k_p} \geq mem_p$. Note that $n^k_1=n^k_P=d_k$ since the first and last part are always assigned to the respective data owner, 
so $mem_{d_k} \geq mem_1 + mem_P$. Subject to suitable partitioning, this is realistic for a wide range of models even for resource-constrained devices. 
Each of the intermediate parts is assigned to a different compute node, thus $n^k_p \neq n^k_{p'}, 2 \leq p \neq p' \leq P-1$, 
but the same intermediate part is hosted on the same 
compute node for all data owners, $n_p = n^k_p = n^{k'}_p, 2 \leq p \leq P-1$.  
Note, however, that the compute node maintains (and trains) an independent instance of the model for each data owner. 

\subsection{Pipeline delay}

Training for each batch of $d_k$ is performed through a bidirectional pipeline involving the nodes that are assigned the different model's parts. Specifically, during the forward propagation phase, $n^k_p$ performs the forward() computation for $M_p$ and sends the activations of the cut layer to $n^k_{p+1}$ which then starts the computation for the next part $M_{p+1}$. Conversely, during the back-propagation phase, $n^k_{p+1}$ computes the backward() for $M_{p+1}$ and sends the respective gradients to $n^k_p$ which then starts the computation for $M_p$. 

Note that the latency of the pipeline in each direction (completion time for all model parts in the pipeline) is defined by the slowest node.  
This is illustrated in Fig.~\ref{fig:pipeline-throughput}, where the training tasks  
are split into two or three parts and are assigned to a corresponding number of compute nodes.  
The latency can then be reduced up to half and respectively up to one-third of the original processing time, provided the task is split evenly among the nodes (left scenarios).  
Otherwise, the latency is determined by the slowest node (middle and right scenarios). 
Thus, the latency of the pipeline for the forward-propagation and backward-propagation can be expressed~as:
\begin{eqnarray}
L_{fwd} = max_{p=2}^{P-1}(procT^{fwd}_p(n_p))   \label{eq:forward} \\
   L_{back} = max_{p=2}^{P-1}(procT^{back}_p(n_p)) \label{eq:backprop} 
\end{eqnarray}
\noindent where $procT^{fwd}_p(n_p)$ and $procT^{back}_p(n_p)$ 
are the processing times 
for the forward() and backward() steps for a single batch  
on the compute node 
responsible for model part $M_p$. 
Then, the average batch processing time when the \textit{pipeline is full} (every node has at least one forward() and one backward() task), is
\begin{eqnarray}
\label{eq:batch}
   procT_{batch}^{pipefull} = L_{fwd} + L_{back}
\end{eqnarray}
We assume that each node performs the training tasks concurrently with the data transmissions. This is reasonable given that modern computing platforms can efficiently overlap computation with communication.

The pipeline is empty when a new global epoch starts.
The time for the very first batch to get processed by the pipeline, and for the pipeline to get filled with training tasks, is
\begin{eqnarray*}
\label{eq:pipeempty}
  procT^{pipeempty}_{batch} = \sum_{p=2}^{P-1}{procT^{fwd}_p(n_p)} + \sum_{p=2}^{P-1}{procT^{back}_p(n_p)}
\end{eqnarray*}

\noindent The equation assumes that the compute nodes are idle when processing 
the forward() and backward() tasks for the first batch. This is true for the forward() tasks, but not for the backward() tasks; when these 
arrive, the pipeline is filled with the forward() tasks. 
For simplicity this contention is ignored, making the equation 
an optimistic lower bound. 

Furthermore, there is an additional delay before the first batch starts being processed by the 
pipeline, which is the time needed by the first data owner to perform the local forward() task for $M_1$ and send the activations to the compute node $n_2$ responsible for model part $M_2$. This delay is estimated as   
\begin{eqnarray*}
\label{eq:batch_first}
startT^{first}_{batch} = \frac{1}{K}\left(\sum_{k=1}^K{procT^{fwd}_1(d_k)} + commT^{fwd}(d_k, n_2) \right)
\end{eqnarray*}
where $commT^{fwd}(d_k,n_2) = \frac{data^{fwd}_{1,2}}{bw_{d_k,n_2}}$ is communication delay for the transfer of the activations and $data^{fwd}_{1,2}$ is the respective amount of data that needs to be transferred between $d_k$ and $n_2$. The reason for using the average overall data owners is that we do not know beforehand which one will send the first batch to the pipeline.

There is a similar delay for the last batch of the epoch after this has been processed by the pipeline, to transfer the gradients from $n_2$ to the last data owner and to perform the local backward()  
for the first model part $M_1$, which is not overlapped by other processing tasks. This delay is equal to
\begin{eqnarray*}
\label{eq:end_batch}
endT^{last}_{batch} = \frac{1}{K}\left(\sum_{k=1}^K{commT^{back}(n_2, d_k)} + procT^{back}_1(d_k)\right)
 \end{eqnarray*}

Like in FL, in Parallel SL, and also in MP-SL, 
all $K$ data owners train the model by feeding each batch $r$ consecutive iterations~(each such iteration corresponds to a so-called \textit{local epoch}). The global epoch is completed by synchronizing the model via aggregation. The total time needed for this is   
\begin{eqnarray}
\label{eq:allepochs}
  T_{batch}^{all} &=& startT^{first}_{batch} + procT^{pipeempty}_{batch} \\ 
  &+& (r \cdot B \cdot K - 1) \cdot procT^{pipefull}_{batch} + endT^{last}_{batch} \notag
\end{eqnarray}

\subsection{Aggregation and global epoch delay}
\label{sec:cost-all}
The aggregation of the model parts corresponding to each data owner is performed \textit{independently} for each of the intermediate parts $M_p, 2 \leq p \leq P-1$ by the compute node $n_p$ responsible for that part. Notably, no communication is required for this between the data owners and/or the compute~nodes.  

However, the aggregation for the first and last parts~(hosted on the data owners), does require extra communication. We assume, similarly to FedAvg, that a designated compute node $c_{aggr}$ is used exclusively for this purpose (the node is not part of the pipeline). When a data owner completes the last batch, it sends the model updates to $c_{aggr}$. Note that, for the large majority of the data owners, this communication takes place while the pipeline is processing other tasks. Therefore, this communication largely overlaps with the training phase and does not introduce any significant additional delay. 
Nevertheless, for $c_{aggr}$ to start the actual aggregation, it needs to wait until it receives the model updates from the last data owner. Let $data^{aggr}_{M_1}$ and $data^{aggr}_{M_P}$ be the amount of data each data owner needs to exchange with $c_{aggr}$ for the first and last part of the model, respectively. Also, let  $commT^{aggr}_{M_1,M_P}(d_k,c_{aggr}) = \frac{data^{aggr}_{M_1} + data^{aggr}_{M_P}}{bw_{c_{aggr}, d_k}}$ be the delay for the data transfer between $d_k$ and $c_{aggr}$. Then, the aggregation delay is equal to
\begin{eqnarray*}
\label{eq:aggr}
  T_{aggr} =  \frac{1}{K}\left(\sum_{k=1}^K(commT^{aggr}_{M_1,M_P}(d_k, c_{aggr}))\right) \notag \\
+ procT^{aggr}(M_1, M_P) +  \sum_{k=1}^K commT^{aggr}_{M_1,M_P}(c_{aggr}, d_k)
\end{eqnarray*}
The first term captures the delay for transmitting the model updates from the last data owner to $c_{aggr}$. Since this could be any of the data owners, the delay is estimated using the average communication delay overall data owners. 
The second term is the time needed by $c_{aggr}$ to aggregate all 
$M_1$ and $M_P$ updates 
and produce the respective global updated parts. Finally, the third term is the delay in the transmission of the updated model parts from $c_{aggr}$ back to the data owners. 

Based on all the above, the total delay for the completion of a global epoch, including aggregation, is equal to

\begin{eqnarray}
\label{eq:global-epoch}
  T_{tot} =  T_{batch}^{all} + T_{aggr}
\end{eqnarray}

\subsection{Training cost model for benchmarks} 

In the same manner, we present an analytical cost model for SplitNN and horizontally scaled Parallel SL. This allows us to fairly compare with MP-SL in the Evaluation section.

\label{sec:bench-cost}
\noindent \textbf{SplitNN.} The model is split into three parts ($P=3$) and it is sequentially trained by each data owner using a single compute node assigned to $M_2$. Once a data owner completes its model updates it sends to the next data owner~(following a round-robin order) the updates of its local model parts; $M_1$ and $M_3$.
The training round is completed when all data owners have updated the model using their 
data. Note there is no notion of a local epoch  
since each batch is processed only once and the data owners apply the updates directly to the global model. So, the (global) epoch delay is:  

\begin{eqnarray*}
\label{eq:end_batch}
T_{tot}^{SplitNN} = \sum_{k=1}^K \left(T_{proc}^{SplitNN} + T_{comm}^{SplitNN} + commT_{M_1, M_3}\right)
 \end{eqnarray*}

\noindent Where, 

\begin{eqnarray*}
T_{proc}^{SplitNN} =  B\cdot\sum_{p=1}^3 \left(procT^{fwd}_p(n_p) + procT^{back}_p(n_p)\right)
 \end{eqnarray*}

\noindent and,

 \begin{eqnarray*}
T_{comm}^{SplitNN} = 2\cdot B\cdot (commT^{fwd} + commT^{back}))
 \end{eqnarray*}

\smallskip \noindent For each batch update, the data owner sends the activations of the first cut layer to the compute node and receives the activations from the second cut layer, and vice versa for the gradients. The last term of the equation is the cost of sending the updated first and last model parts to the next data owner. 

\smallskip \noindent \textbf{Horizontally Scaled Parallel SL.} The model is split into three model parts~($P=3$). Each compute node is in charge of a different set of data owners. 
Primarily, we make sure that the data owners are proportionally distributed among them; taking into account the computing and memory capacity of the compute nodes.
The training delay is determined by the compute node which takes the longest time to complete.

After assignment, the delay for each compute node can be computed independently using the Eq.~\ref{eq:allepochs}. The delay of the training phase is equal to the delay of the compute node finishing last.
Note, however, that there is an additional cost in the aggregation phase. As all compute nodes need to synchronize with each other the model parts they handle. Hence, they will, as well, send to the aggregator their model updates of $M_2$.
\section{Split selection}
\label{sec:split}

Given the sets of $K$ data owners, $N$ compute nodes~(i.e, $N=P-2$), and the first/last cut layers,  
the split points and assignment of the intermediate model parts~(i.e, the ones offloaded to the compute nodes) can be optimized, by minimizing the latency of the full-pipeline~(Eq.~\ref{eq:batch}). 

To do so, we formulate a problem with decision variable $ x=(x_{ij} \in \{0,1\}, (i\in N, j\!\in\! S^*))$, where $S^* \subseteq S$ is the subset of the layers that comprise the intermediate model part; does not contain the layers before the first and after the last cut. The $x_{ij}=1$ if layer $j$ is offloaded to compute node $i$. But, recall that each layer can only be offloaded into one compute node, and each compute node receives at least one layer,
\begin{eqnarray}
\label{eq:c1}
\sum_{i=1}^N x_{ij} = 1, \textrm{ } \forall j \in S^* \textrm{ and } \sum_{j=1}^{S^*} x_{ij} \geq 1, \textrm{ } \forall i \in N
 \end{eqnarray}
The compute nodes handle model parts with 
sequent layers,
\begin{eqnarray}
\label{eq:c2}
\sum_{k=2}^{j} x_{ij}(x_{i,k-1}-2x_{ij}-x_{ik}) \leq 0,\textrm{ }\forall j \in S^* \textrm{, }\forall i \in N 
 \end{eqnarray}
Also, recall that $mem_{c_i}$ is the available memory of compute node $c_i$. Let $d_j$ be the memory demand for layer $j$, hence
\begin{eqnarray}
\label{eq:c3}
K \sum_{j=1}^{S^*}x_{ij} d_j\leq mem_{c_i}, \textrm{ } \forall i \in N
\end{eqnarray}
Then, the equations~\ref{eq:forward} and \ref{eq:backprop} can be updated accordingly,
\begin{eqnarray}
L_{fwd} = max_{p=2}^{P-1}(\sum_{j=1}^{S^*} x_{pj}procT^{fwd}_j(n_p))
\label{eq:c4} \\
L_{back} = max_{p=2}^{P-1}(\sum_{j=1}^{S^*} x_{pj}procT^{back}_j(n_p)) \label{eq:c5} 
\end{eqnarray}
Finally, the objective is to find the splitting policy, that minimizes the pipeline delay:
\begin{eqnarray}
\mathbb{P}:
\minimize_{x,procT,L} procT_{batch}^{pipefull}    \\
\text{s.t.} \text{ }(\ref{eq:batch}), (\ref{eq:c1})-(\ref{eq:c5})
\label{eq:objective} 
\end{eqnarray}

\begin{figure}[t]
    \centering
        \includegraphics[width=\columnwidth]{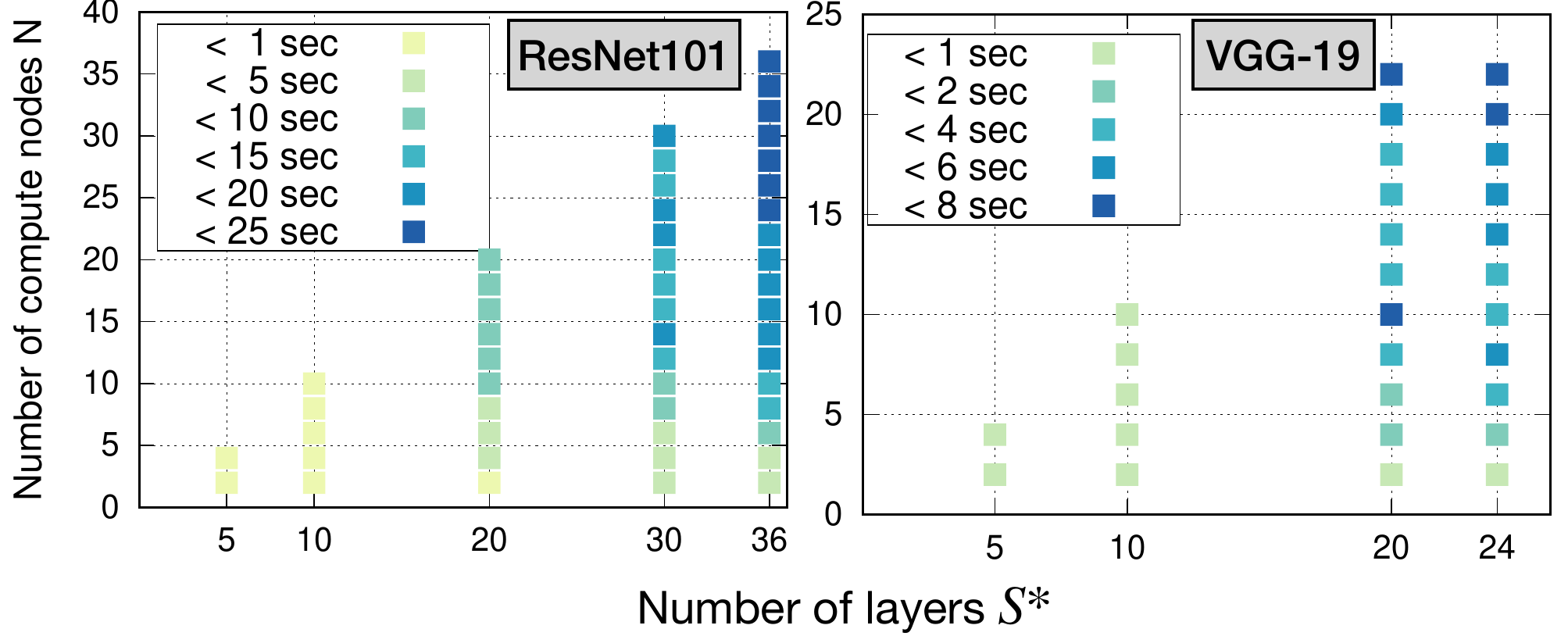}
    \caption{Computing time for optimizing the problem with Gurobi, while the size of the problem changes.
    \vspace{-0.3cm}}
    \label{fig:solver}
\end{figure}

\begin{figure*}[t] 
    \centering
    \includegraphics[clip,width=1.85\columnwidth]{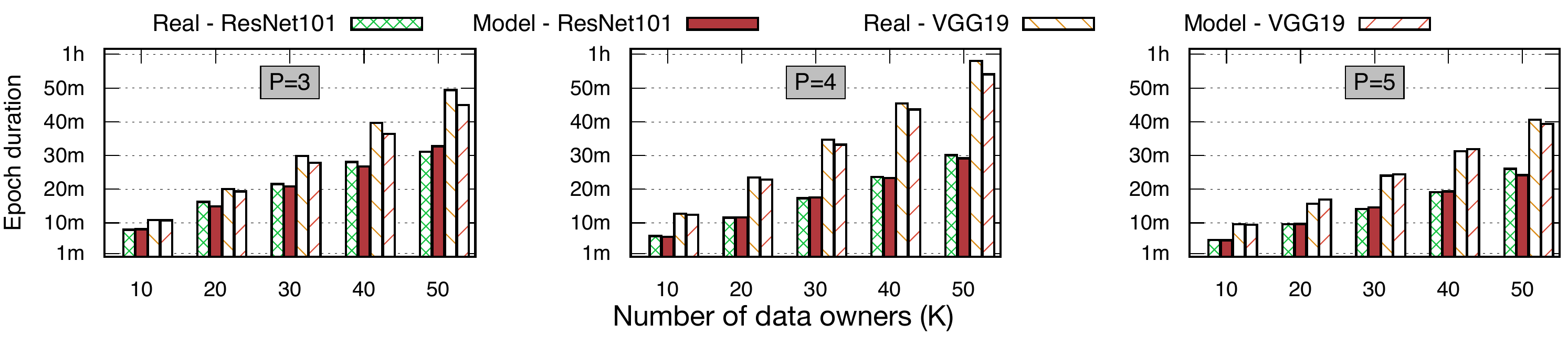}
    \caption{Average training performance of real experiments for $d_1$ vs the performance estimated by the model.\vspace{-0.3cm}}
    \label{fig:model_check}
\end{figure*}

MP-SL uses Gurobi~\cite{gurobi}, a well-known ILP solver, to solve problem $\mathbb{P}$. As we will show the additional overhead of optimizing the splitting decision is negligible considering the acceleration gain. In  Fig.~\ref{fig:solver} we conduct a sensitivity analysis of the problem's size; the dimension of $x$ variable. This is proportional to the number of possible splits $S^*$ and the number of compute nodes $N$. Fig.~\ref{fig:solver} shows the computing time of the solver as the values of $S^*$ and $N$ change. Note, that the first/last cut layers determine $S^*$, and hence we alter these split points accordingly for each experiment.
We notice that the computing time is more sensitive to the number of compute nodes. This can be seen in the left plot~(i.e., using ResNet-101) when $S^*$ has the largest values, the computing time gets greater as $N$ ascends. We notice the same effect for the VGG-19 model~(right plot). Nevertheless, the optimizing cost remains negligible, considering that this is a one-time overhead since the manager of MP-SL will only optimize $\mathbb{P}$ during the offline period before training starts. Also, in the Evaluation section, we will study the acceleration gain and show that this overhead can be ignored.
\begin{table}[t]
    \centering
    \begin{tabular}{p{0.2cm}p{1.2cm}p{2.4cm}p{0.9cm}p{0.7cm}p{0.7cm}}
    \toprule
    ID & 
    Platform & CPU & Memory & ResNet & VGG\\ \cmidrule{1-6}
    $d_1$ & \footnotesize RPi~4 B & \footnotesize Cortex-A72  (4 cores) & \footnotesize 4GB &  \footnotesize $91.9(1.8)$ & \footnotesize $71.9(1)$
    \\
    $d_2$ & \footnotesize RPi~3 B+ & \footnotesize Cortex-A5 (4 cores) & \footnotesize 1GB &  \multicolumn{2}{c}{\footnotesize no memory}\\
   \footnotesize VM & \footnotesize CentOS~7.9 & \footnotesize 8-core virtual CPU & \footnotesize 16GB & \footnotesize $2 (0.18)$ & \footnotesize $3.6(0.1)$\\ 
    \bottomrule
    \end{tabular}
    \caption{Testbed nodes and average computing time in seconds~(standard deviation in brackets) for a batch update, where the batch size is $128$ samples. 
    \vspace{-0.3cm}}
    \label{table:physica-device}
\end{table}

\section{Evaluation}
\label{sec:evaluation}

In this section, we present the evaluation of MP-SL. As indicative ML models, we use ResNet-101~\cite{he2016deep} and VGG-19~\cite{simonyan2014very} trained with the CIFAR-10~\cite{krizhevsky2009learning} dataset.  
Considering the high interconnectivity between the layers in residual blocks in ResNet, we consider them as potential atomic unit-blocks. While, in VGG, every single layer is an atomic unit-block. The models are trained using $16$ batches~($B=16$) of $128$ samples.
Also, we assume that there are $r=2$ local epochs before performing the aggregation step to complete a global epoch. 
Notably, the selection of $B$ and $r$ is arbitrary as they are hyper-parameters of the training procedure while MP-SL is designed to enable resource-constrained devices to participate in FL and reduce the duration of a training epoch.

\smallskip \noindent \textbf{Testbed.} We use 
a physical testbed to measure the performance of MP-SL. 
For the role of data owners, we use two different Raspberry Pi devices. For the compute nodes and the manager, we use VMs running in a private cluster. The communication between data owners and compute nodes is via VPN over WiFi and the public Internet. The hardware characteristics of the data owner devices and compute nodes are given in Table~\ref{table:physica-device}, which also lists the average time needed to perform one batch update for the full ResNet-101 and VGG-19 models. Note that $d_2$ cannot support on-device training due to memory limitations, 
thus such devices can't participate in FL. Also, even though $d_1$ can support on-device training, this takes a very long time due to its limited computing capabilities, which would cause a straggler's effect if $d_1$ were to participate in FL with faster devices. It is precisely in such cases that MP-SL can enable such devices to participate in collaborative training efforts.
 
\smallskip 
\noindent \textbf{Emulation of numerous data owners.}
Since we only have a few data owner devices at our disposal, it is not possible to perform large-scale experiments. Instead, we emulate a large number of data owners using additional nodes in our cluster. 
To this end, we profile the forward() and backward() tasks on $d_1$ and $d_2$ for ResNet-101 and VGG-19 
and measure the throughput between $d_1$ and $d_2$ and the VMs. These measurements are then used to run multiple data owner processes on 
the VMs of the cluster. To mimic the behavior of the resource-constrained devices, processing and data transfers are artificially slowed down as needed. When the VM reaches its processing capacity, we add another one to emulate the rest of the data owners.  
For VGG-19, whose memory demands exceed the available memory on the VMs when running numerous emulated data owners, we remove the last two layers, 
while keeping the processing and transferring delays the same as for the full model part. 
Note that the (emulated) data owners and the compute nodes run the full MP-SL framework, exactly as done in system configurations where the real $d_1$ and $d_2$ devices are used with the compute nodes in the~cluster.

\subsection{Model validation}

In the first set of experiments, we measure the epoch time of MP-SL and compare it with the estimate obtained via our model (i.e., the output of Eq.~\ref{eq:global-epoch}). Fig.~\ref{fig:model_check} shows the results with one ($P=3$), two ($P=4$) and three ($P=5$) compute nodes, 
for $10$ up to $50$ data owners of type $d_1$. 
As can be seen, the model is close to the real results, with an average absolute error of $3.14\%$ over all experiments for ResNet-101 and $3.86\%$ for VGG-19. 
Note, however, that when there are 
fewer splits and, as a result, the compute nodes 
are assigned larger model parts, the delay estimation 
of the model is smaller than the 
actual measurements. 
There is a similar deviation as the number of data owners 
increases. This is a side-effect of the memory 
pressure in the compute nodes, which is not captured by the model. Nevertheless, the model is sufficiently accurate to serve as a tool for investigating a wide range of scenarios without having to run real experiments.

\begin{figure}[t]
    \centering
        \includegraphics[width=0.95\columnwidth]{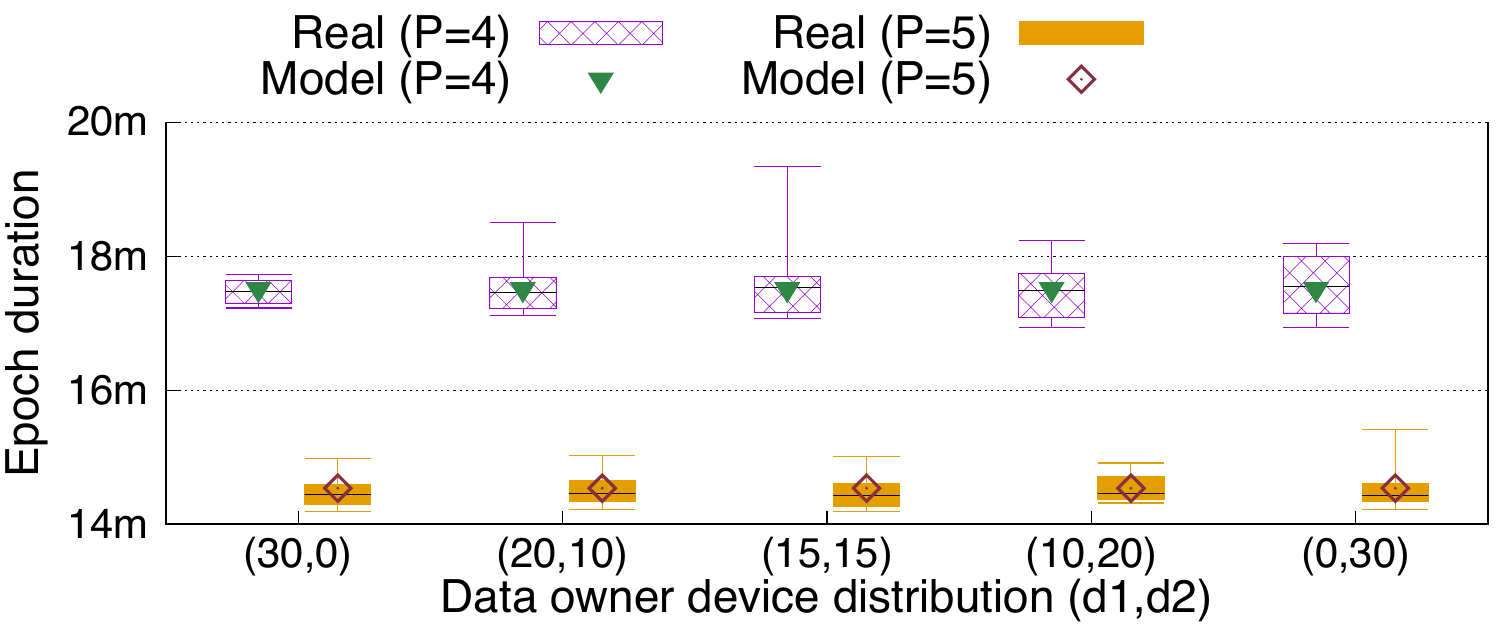}
    \caption{Model validation for heterogeneous data owner devices, with $30$ data owners, two ($P=4$) and three ($P=5$) compute nodes. 
    The distribution  
    notation $(q_1, q_2)$ 
    means that there are  
    $q_1$ devices of type $d_1$ and 
    $q_2$ devices of type $d_2$. \vspace{-0.3cm}}
    \label{fig:heterog}
\end{figure}

Furthermore, we run experiments with a heterogeneous distribution of data owners and compare the measured 
epoch delay with the estimates of our model. Fig.~\ref{fig:heterog} shows the results of training ResNet-101 with two and three compute nodes for $30$ data owners, as the portion of  
$d_1$ vs $d_2$ devices varies. A single batch of 
$d_2$ needs roughly $7.25$~seconds to complete the forward() and back() tasks of the first and last model part, while $d_1$ 
needs about $3.15$~seconds for this, thus is $2.3x$ faster. We observe that the data owner's characteristics do not affect the training time significantly. For instance, the time difference for the faster case~(only $d_1$) to the slowest case~(only $d_2$) is merely $0.08s$ for three compute nodes and just $0.062s$ for two compute nodes. This is reasonable because 
the duration of the (global) epoch is 
affected by the data owner's characteristics merely in the first and last batch~(Eq.~\ref{eq:global-epoch}).

The median of the real epoch measurements has 
a small fluctuation, 
up to $1.98$ seconds for three compute nodes ($P=5$) and $5$ seconds for two ($P=4$). Also, it is very close to the model estimates.
Note that, for the same system configuration,  there is some variance in the measurements, with a few outliers. This is a side effect of running experiments on a shared infrastructure like our cluster or in the cloud, where VMs can occasionally experience a slowdown in their execution due to multi-tenancy.

\begin{figure}[t]
    \centering
    \includegraphics[width=0.95\columnwidth]{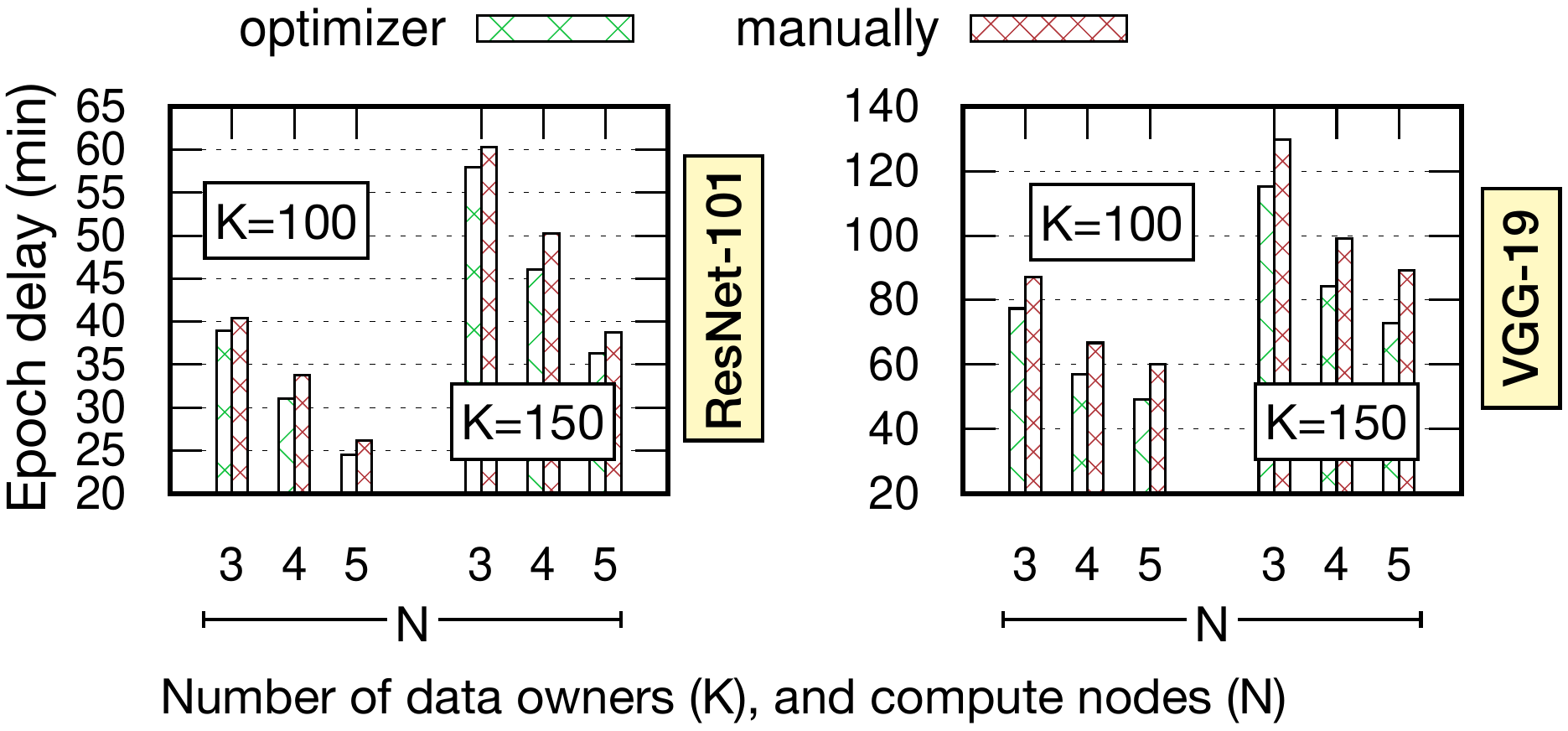}
    \caption{Epoch duration when split points are optimized and manually selected, for different scenarios of $K$ and $N$.\vspace{-0.4cm}}
    \label{fig:split}
\end{figure}

\subsection{MP-SL Exploration}

After validating the performance estimation model~(Eq.~\ref{eq:global-epoch}), we explore the performance of MP-SL for more nodes.

\smallskip
\noindent \textbf{The importance of splitting optimization.}
We evaluate the importance of optimizing splitting for the intermediate model parts. MP-SL optimizes the selection of the intermediate split points (Sec.~\ref{sec:split}), using an ILP solver with an additional overhead less than $1$ sec for (up) to 5 compute nodes~(Fig.~\ref{fig:solver}). This is negligible compared to the typical time required to train a model for multiple epochs, ranging from minutes to~hours.

Fig.~\ref{fig:split} compares the duration of one epoch
when using the splitting mechanism of MP-SL versus a self-designed benchmark approach,\footnote{A direct comparison to another research work is not feasible, as to the best of our knowledge there are no works for multihop Parallel SL~(Sec.~\ref{sec:background}).} 
where the split points are selected manually. Specifically, we divide the total computing time of the intermediate part by the number of compute nodes. This way, we assume a pipeline latency that could evenly distribute the computing cost.
Then, we manually assign to the compute nodes subsequent layers that sum up to a latency close to the computed one~(i.e., trying not to exceed it).\footnote{In case of heterogeneous compute nodes this approach is not as trivial. Unlike MP-SL which optimizes splitting for any case.} However, this is not always feasible, as the computing demands for each layer can be arbitrary and do not have any inherited symmetry~\cite{kang2017neurosurgeon}. As is shown in Fig.~\ref{fig:split} using the framework's optimizer, the improvements per-epoch can be up to $8.5\%$ for the ResNet-101 and up to $19\%$ for the VGG-19. The VGG-19 is a more challenging model to manage manually as the per-layer computing cost varies more than the one in ResNet-101.

\begin{figure}[t]
    \centering
    \includegraphics[width=\columnwidth]{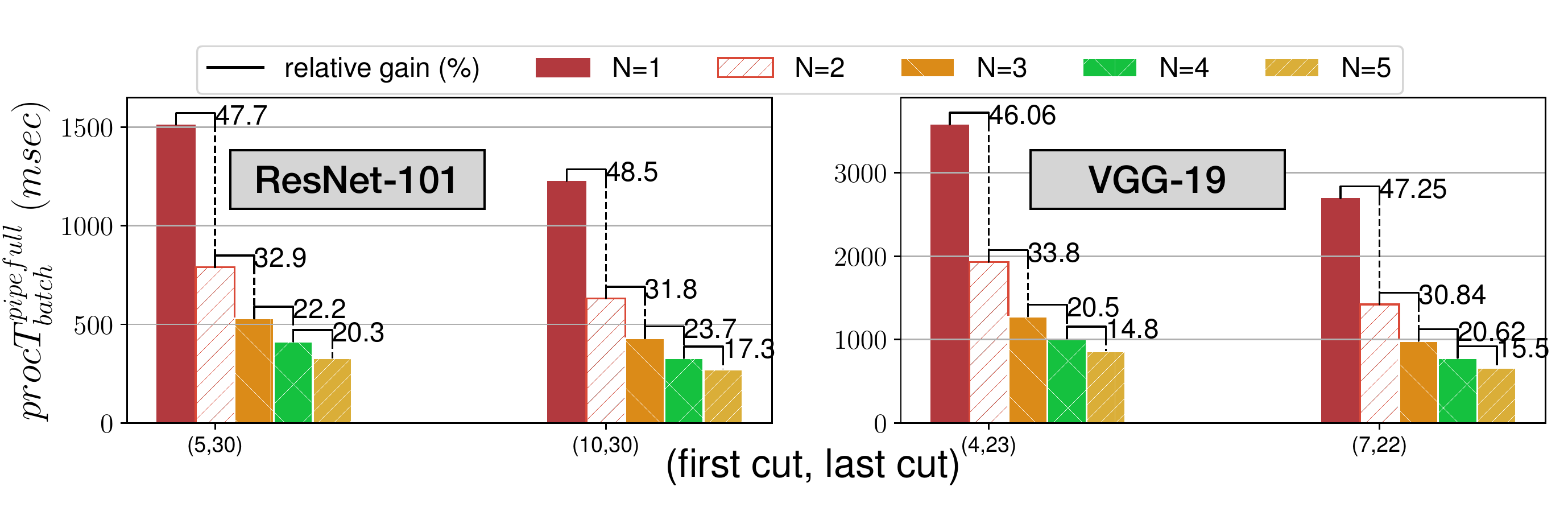}
    \caption{Pipeline's latency as the multihop level increases. 
    \vspace{-0.3cm}}
    \label{fig:dimish}
\end{figure}

\begin{figure}[t]
\begin{center}
\subfloat[Heterogeneity in data owners' devices. The parenthesis has the number of $d_1$ and $d_2$ devices respectively.]{
  \includegraphics[width=0.80\columnwidth]{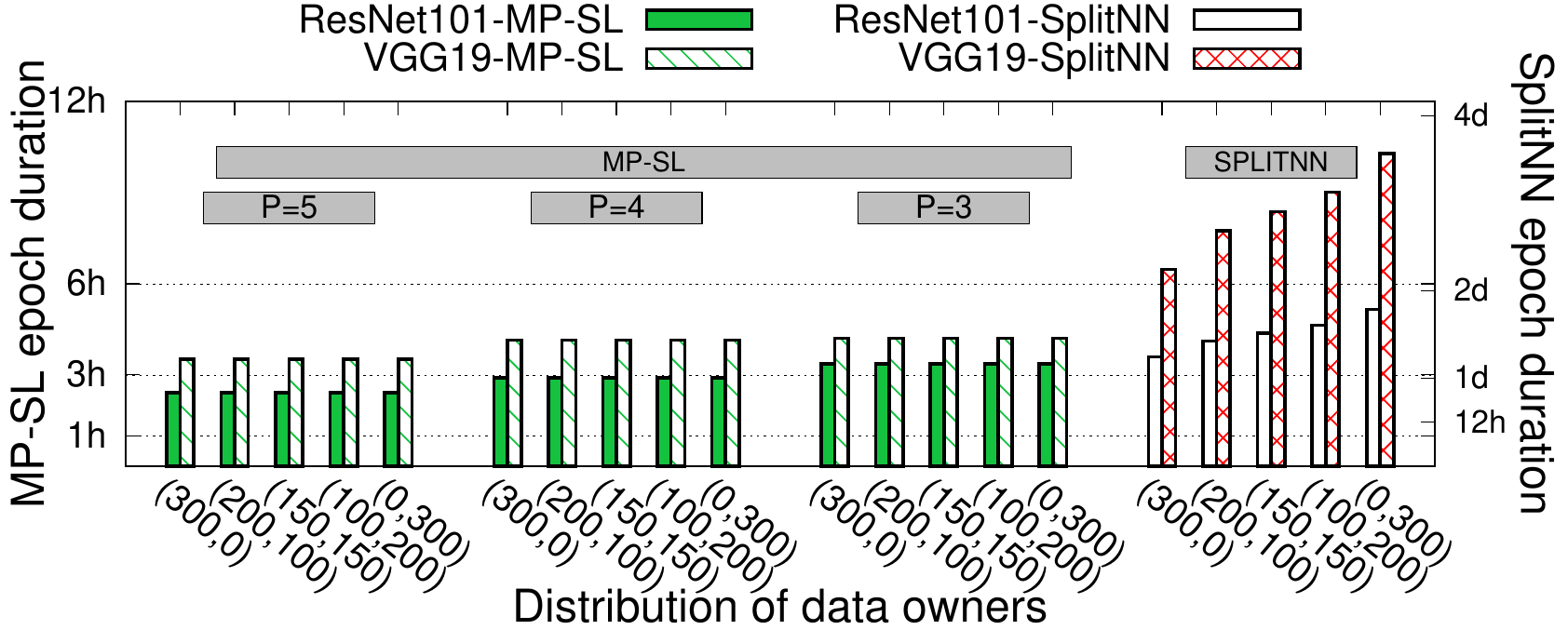}
  \label{fig:heter}
} \\
\subfloat[Heterogeneity in the data owners' network. There are 300 data owners in total.]{
  \includegraphics[width=0.80\columnwidth]{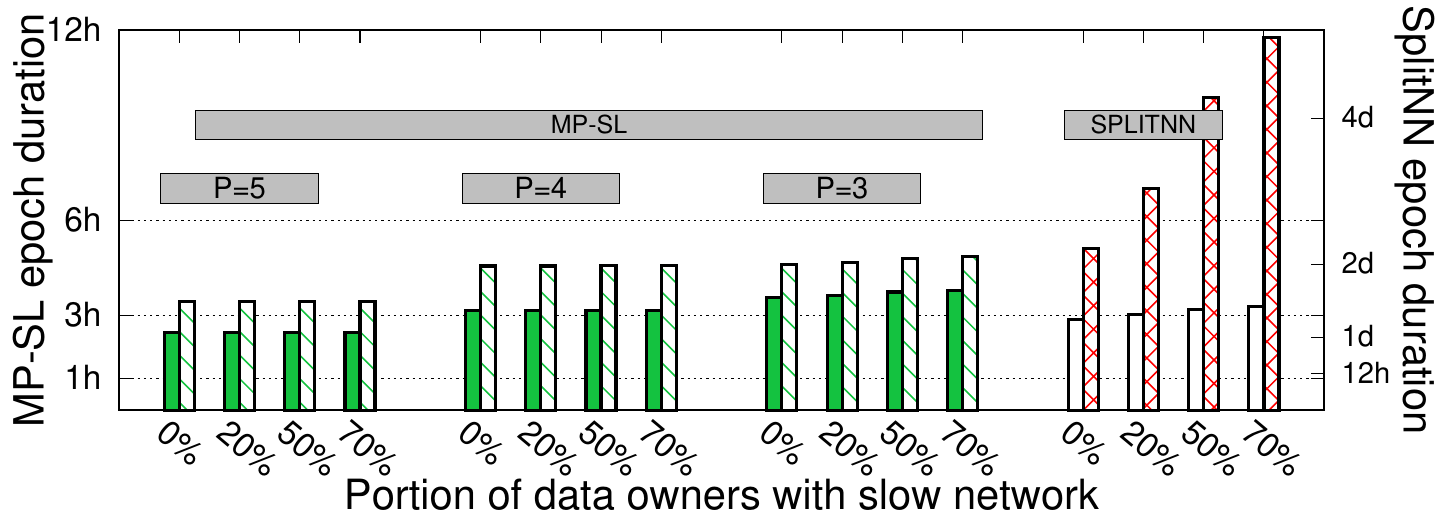}
  \label{fig:heter_net}
}
\end{center}
\vspace{-0.2cm}
\caption{Epoch delay for heterogeneous cases using MP-SL with one ($P=3$), two ($P=4$), and three ($P=5$) compute nodes vs SplitNN (second y-axis). Both y-axis are in log-scale.\vspace{-0.3cm}}
\label{fig:heter_all}
\end{figure}

\begin{figure*}[ht]
    \centering
    \includegraphics[width=1.85\columnwidth]{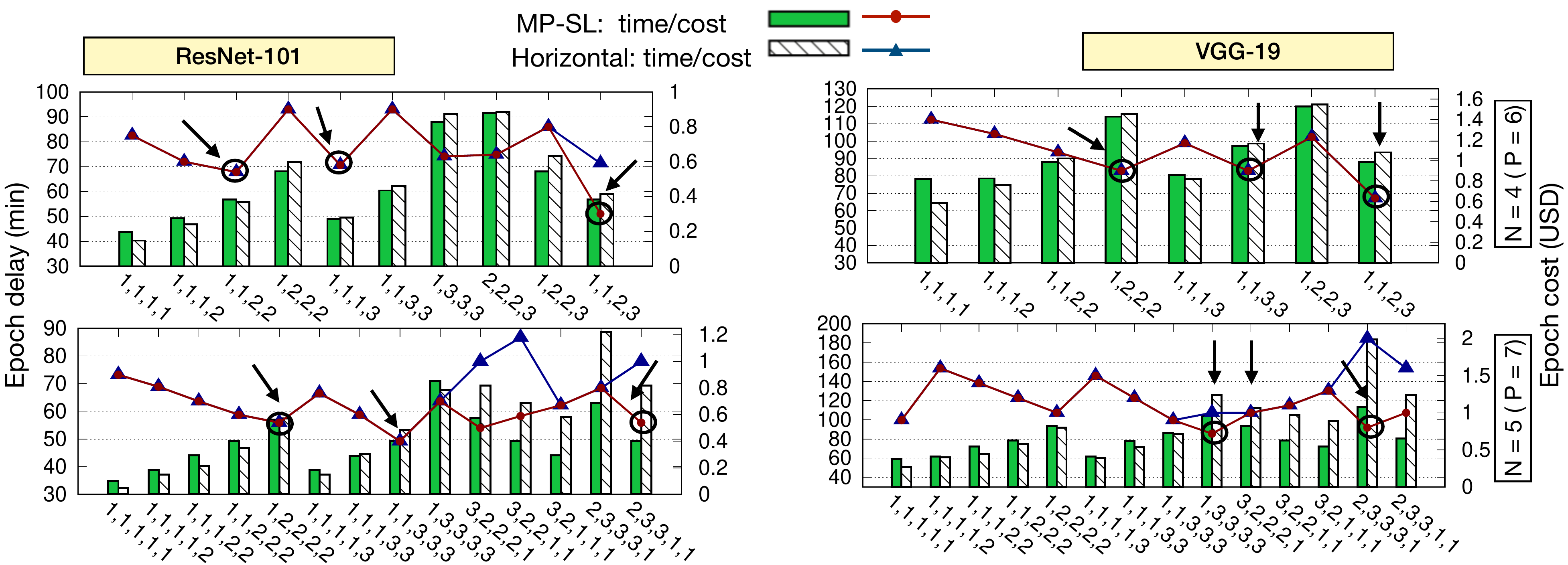}
    \caption{Epoch delay and cost for MP-SL and Parallel SL horizontally scaled configuration for scenarios with $150$ data owners in collaboration with four~(top) and five~(bottom) compute nodes. 
    The x-axis shows the ID of the VM instance type.
    \vspace{-0.2cm}
    }
    \label{fig:horizontal}
\end{figure*}

\smallskip \noindent \textbf{Different multihop level.} Fig.~\ref{fig:dimish} shows the latency of the pipeline (Eq.~\ref{eq:batch}) as more compute nodes are added, for different first/last cut layers. It is obvious that as the multihop level increases the latency of the pipeline decreases, significantly. For instance, by adding one compute node to the smallest possible multihop level~($N=1$ or $P=3$), the improvement can be up to $46\%$. Finally, as expected, the delay decreases at a smaller pace as the multihop level gets larger.

\smallskip
\noindent \textbf{MP-SL in a heterogeneous system.}
One of the main challenges in distributed learning is the straggler effect. As we will show, even SplitNN is very sensitive to that, yielding significantly higher epoch delays when slower (computing and network-wise) data owners participate in the training procedure.
Whereas, MP-SL is independent of the data owner's characteristics when the pipeline is fully utilized.

This is shown in Fig.~\ref{fig:heter}, 
for scenarios with a total of $300$ data owners where we vary the portion of  $d_1$ vs $d_2$ devices. We examine how the epoch duration changes in these heterogeneity scenarios for MP-SL with one, two, and three compute nodes 
vs SplitNN.~\footnote{Estimated using the analytical model described in Sec.~\ref{sec:bench-cost}}
The epoch delay in SplitNN 
for $300$ $d_2$ devices is $30\%$ higher than the one with $300$ $d_1$
devices. In contrast, in MP-SL with two or three compute nodes the difference is negligible, confirming that slower data owners do not harm the training performance. 

Also, in Fig.~\ref{fig:heter_net} we study the impact of training in a network heterogeneity context. It shows the performance for scenarios with a total of $300$ data owners as we vary the portion of data owners that have a slower network connection~(i.e., up to $8$ times slower than the profiled throughput). The epoch delay of SplitNN when $70\%$ of the data owners have a slow network connection is $12\%$, and $56\%$ higher for ResNet-101, and VGG-19, respectively, compared to the case where there is no slow network connection. Namely, VGG is more sensitive to network characteristics because it is a heavier model with many more parameters than ResNet. 
Whereas, in MP-SL there was no change in performance for $P=5,4$ and a small increase (up to $6\%$) for $P=1$ in the case of $70\%$. 

\smallskip
\noindent \textbf{Cost \& training time of MP-SL vs. horizontally scaled Parallel SL.} Inspired by the Data Parallelism~(DP)~\cite{huang2019gpipe}, Parallel SL can be extended to support horizontal scaling, in which 
several compute nodes are in charge of the whole intermediate model part, but assist different data owners. The data owners are allocated to the compute nodes proportionally, depending on the computing capacity of each compute node. Horizontal scaling can achieve remarkable acceleration.  
However, to accomplish that one should have access to VMs~(to run the processes of the compute nodes) with sufficient memory and computing resources. Notably, this increases significantly the cost when the computing resources are employed in a pay-as-you-go model as the price of a VM depends on its characteristics.
For instance, Table~\ref{table:aws_same_mem} shows three indicative types of AWS instances: (i)~$vm_1$ has the highest price but is the most powerful one, (ii)~$vm_2$, with half the price, has half memory and vCPUs, and (iii)~$vm_3$, with $1/4$ of $vm_1$'s price, has the same vCPUs as $vm_2$, but half memory size. 
Thus, the selection of the VMs  
creates a \textit{cost-delay trade-off}. 

We study the effect of VM selection, 
by constructing a similar set of VMs. Considering that $vm_1$ has equal computing capacity as the compute nodes we have profiled~(Table~\ref{table:physica-device}), while $vm_2$ and $vm_3$ are two times slower than the profiled data; they have half as much vCPUs. Also, the memory of the VMs is accordingly defined.
We input this data into the analytical models of MP-SL~(Sec.~\ref{sec:cost-all}) and  
horizontally scaled Parallel SL~(Sec. \ref{sec:bench-cost}) to estimate the epoch's delay. Using this analogy we estimate the cost of the epoch. 
We consider four different scenarios --training ResNet-101 and VGG-19 with four and five compute nodes-- and, as is shown in Fig.~\ref{fig:horizontal}, we compute the epoch's delay and cost while altering the combination of the instances type from the set of VMs. 

\begin{table}[t!]
    \centering
    \begin{tabular}{p{0.3cm}p{0.95cm}p{0.5cm}
    >{\centering\arraybackslash}p{1.7cm}>{\centering\arraybackslash}p{2.5cm}}
    \toprule
     ID & Instance & vCPUs & Memory (GB) & Cost per hour (USD)\\ \hline
     $vm_1$ & t2.xlarge & 4 &  16 & 0.18 \\ 
     $vm_2$ & t2.large  & 2 &  8  & 0.092 \\ 
     $vm_3$   &  t2.medium & 2 & 4 & 0.046 \\
    \bottomrule
    \end{tabular}
    \caption{Data derived from the cost catalog of AWS.
    \vspace{-0.3cm}}
    \label{table:aws_same_mem}
\end{table}

Firstly, examining individually each case of VM selection, we notice that in most cases MP-SL outperforms the horizontal scaling configuration when using the cheaper compute nodes~(i.e., $vm_2$, $vm_3$). Horizontally scaled Parallel SL can be $\times 1.4$ and $\times 1.62$ slower than MP-SL when using five compute nodes to train ResNet-101 and VGG-19, respectively. Moreover, 
for the same VM selection, the cost can increase up to $50\%$ for ResNet-101, and $60\%$ for VGG-19 when using horizontal scaling instead of MP-SL.
However, even when horizontal scaling outperforms MP-SL the relative delay 
do not exceed $\times 1.02$ for ResNet-101 and $\times 1.3$ for VGG-19, and their cost does not differ, as well.

The combination of the VMs with only $vm_1$ achieves the shortest training delay. But this is one of the most expensive solutions. 
Observe that for each scenario in Fig.~\ref{fig:horizontal}
we have marked with arrows the top-3 cases whose VM selection costs the least. 
In all but one marked case, the preferred configuration is MP-SL. For instance, when training ResNet-101 with $P=6$, selecting nodes 1-1-1-3 the cost is $22\%$ less than the only $vm_1$ selection, while the training time gets only $\times 1.2$ slower. MP-SL can reduce the cost, even more,~(up to $60\%$) if we use VMs 1-1-2-3, but the training delay gets $\times 1.4$ slower. Similarly in VGG-19, when selecting nodes 1-1-2-3 the cost is reduced up to $55\%$ with $\times 1.4$ delay. Also, in the same manner for $P=7$, where the cost is dropped up to $55\%$ in ResNet-101 when using the 1-1-3-3-3 node selection, with a delay increase equal to $\times 1.5$. Hence, MP-SL is a framework that can significantly reduce the cost of training with a small increase in the delay.
\section{Conclusions}
\label{sec:conclustions}

In this work, we presented MP-SL a framework that supports distributed and collaborative learning via 
an asynchronous multihop SL protocol. 
Also, we designed a model to estimate the expected performance of MP-SL, with an error smaller than $3.86\%$.  We have shown that the pipeline protocol of MP-SL is robust to heterogeneous systems; often found in distributed learning systems. Finally, an important attribute of MP-SL is that it can perform efficiently even when we use less powerful (cheaper) compute nodes. A natural direction for future work would be the combination of pipeline parallelism and horizontal scaling. To exploit both benefits of the two configurations~(i.e., time acceleration and cost reduction).

\section{Acknowledgements}
This work has been supported by (i) the Science Foundation Ireland (SFI) and the Department of Agricultural, Food and Marine of the Government of Ireland under the Grant Number [16/RC/3835] - VistaMilk, and (ii) the Horizon Europe research and innovation program of the European Union, under grant agreement no 101092912, project MLSysOps.

\bibliographystyle{plain}
\bibliography{references.bib}

\end{document}